\pdfoutput=1

\documentclass[11pt]{article}

\usepackage[preprint]{acl}

\usepackage{times}
\usepackage{latexsym}

\usepackage[T1]{fontenc}

\usepackage[utf8]{inputenc}

\usepackage{microtype}

\usepackage{inconsolata}

\usepackage{graphicx}
\usepackage[table]{xcolor}
\usepackage{booktabs}
\usepackage{multirow}
\usepackage{enumitem}
\usepackage{amsfonts,amsmath,amstext}
\usepackage[caption=false]{subfig}
\usepackage[misc]{ifsym}

\usepackage[many]{tcolorbox}    	
\tcbuselibrary{listings, breakable}
\newtcolorbox{promptBox}[1]{
    title=\textbf{#1},
    breakable,
    fonttitle=\bfseries,
    boxrule = 1pt,
    toprule = 3pt, 
    colframe = teal,
    enhanced,
    rounded corners,
    arc = 2pt,   
    top=1mm,bottom=1mm,left=1mm,right=1mm    
}

\newcommand{\revise}[1]{\textcolor{black}{#1}}
\newcommand{\add}[1]{\textcolor{black}{#1}}

\newcommand{\algname}{\emph{MONAQ}}

\title{\protect\includegraphics[height=16pt]{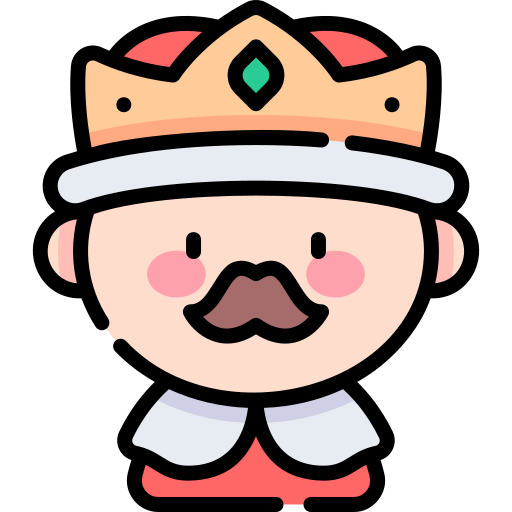} \algname{}: Multi-Objective Neural Architecture Querying for Time-Series Analysis on Resource-Constrained Devices}

\author{Patara Trirat \\
  DeepAuto.ai \\
  \texttt{patara@deepauto.ai} \And
  Jae-Gil Lee \textsuperscript{\Letter} \\
  KAIST \\
  \texttt{jaegil@kaist.ac.kr}
}

\begin{document}
\maketitle
\begin{abstract}

\revise{The growing use of smartphones and IoT devices necessitates efficient time-series analysis on resource-constrained hardware, which is critical for sensing applications such as human activity recognition and air quality prediction. Recent efforts in hardware-aware neural architecture search\,(NAS) automate architecture discovery for specific platforms; however, none focus on general time-series analysis with edge deployment. Leveraging the problem-solving and reasoning capabilities of large language models\,(LLM), we propose \textbf{\algname{}}, a novel framework that reformulates NAS into \textbf{\emph{M}}ulti-\textbf{\emph{O}}bjective \textbf{\emph{N}}eural \textbf{\emph{A}}rchitecture \textbf{\emph{Q}}uerying tasks. \algname{} is equipped with \emph{multimodal query generation} for processing multimodal time-series inputs and hardware constraints, alongside an \emph{LLM agent-based multi-objective search} to achieve deployment-ready models via code generation. By integrating numerical data, time-series images, and textual descriptions, \algname{} improves an LLM's understanding of time-series data. Experiments on fifteen datasets demonstrate that \algname{}-discovered models outperform both handcrafted models and NAS baselines while being more efficient.
}
\end{abstract}

\section{Introduction} \label{section:introduction}
The widespread adoption of smartphones, IoT devices, and wearables has intensified the demand for efficient time-series analysis on \emph{resource-constrained} devices, essential for smart manufacturing, personalized healthcare\,\citep{TinyML_TSC_2024}, and so on. These devices, often based on microcontroller units\,(MCU), are rapidly proliferating, with over 250B units worldwide\,\citep{MCUNet_NeurIPS_2020, MCUNetV2_NeurIPS_2021}. Deep learning on such affordable, energy-efficient hardware can democratize AI, enabling broad accessibility across diverse sectors.

However, tiny deep learning faces unique challenges due to stringent memory constraints. Typical MCUs, with less than 512kB SRAM, and even higher-end devices like Raspberry Pi 4 struggle to run conventional deep neural networks\,\citep{MCUNetV3_NeurIPS_2022}. Efficient AI inference demands innovative methods to navigate these limitations. Moreover, designing optimal network architectures and selecting hyperparameters for such devices is time-consuming and requires significant manual effort. Hardware-aware neural architecture search\,(HW-NAS) has emerged to automate this process, tailoring architectures to specific tasks and hardware\,\citep{HW_NAS_Bench_ICLR_2021, HWNAS_Survey_IJCAI_2021}.

While NAS has advanced, it predominantly focuses on computer vision tasks and lacks generalizability for time-series analysis\,\citep{NAS_1000_Insights}. Time-series applications span classification (e.g., human activity recognition)\,\citep{MLPHAR_ISWC_2024}, regression (e.g., environmental monitoring)\,\citep{TSER_Benchmark_2021}, and anomaly detection (e.g., industrial systems)\,\citep{TimeCAD}. Existing NAS frameworks often target narrow use cases and fail to operate effectively within edge device constraints\,\citep{AutoPytorch_TS_PKDD_2022, PASTA, TinyTNAS_2024}.

Furthermore, current HW-NAS frameworks are frustrated by fixed search spaces and complex interfaces, making them less generalizable across tasks and inaccessible to non-experts. Large language models\,(LLM)-based NAS frameworks\,\citep{automl_llm_survey_TMLR_2024} aim to address these issues but still rely on predefined search spaces and require user-provided initial architectures, limiting flexibility and usability\,\citep{EvoPrompting_NeurIPS_2023, GHGNAS_2023, LLMPP_ACL_2024, LLMatic_2024_GECCO}.

\begin{figure*}[!t]
    \centering
    \subfloat[Existing LLM-based NAS.]{
        \includegraphics[width=0.45\textwidth]{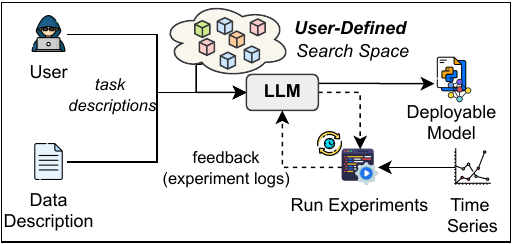}%
        \label{figure:existing_llm_nas}
    }
    \subfloat[Our \algname{}.]{
        \includegraphics[width=0.45\textwidth]{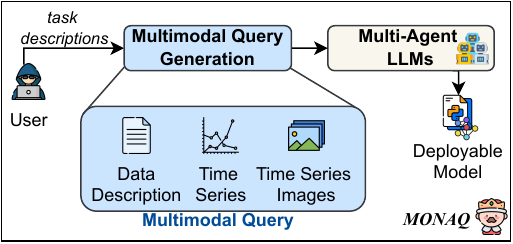}%
        \label{figure:monaq}
    }
\vspace*{-0.2cm}
\caption{\revise{Comparison between (a) existing LLM-based NAS and (b) our proposed \algname{} framework.}} \label{figure:compare_existing_monaq}
\vspace*{-0.5cm}
\end{figure*}

In contrast, to eliminate these undesirable user burdens, we reformulate the NAS problem as a multi-objective \emph{neural architecture querying}\,(NAQ) problem by leveraging LLMs' advancements in reasoning and problem-solving. Unlike existing LLM-based NAS frameworks (Figure~\ref{figure:existing_llm_nas}), which require users to define a search space or an initial set of architectures, NAQ only requires natural language queries from users, decoupling them from the architecture search process. This approach reduces reliance on human expertise while enhancing accessibility and flexibility in the model design process. It allows users to focus on addressing the actual problems in their datasets, leaving the complex search configurations to the LLM. However, achieving efficient NAQ poses two key challenges.

\emph{\textbf{(1) How to find high-performing architectures without user-defined search spaces?}} Without user-defined configurations, enhancing the capability of LLMs to directly design high-performing network architectures becomes crucial. Recent studies\,\citep{MetaGPT_ICLR_24, AgentRisePotential} suggest that multi-agent LLMs improve problem-solving through collaboration among agents specialized in different tasks, while mitigating limitations found in single-agent LLMs, such as bias and hallucination. Building on these insights, we address this challenge by introducing an \emph{LLM agent-based multi-objective search} module (Figure~\ref{figure:monaq}). However, having multiple agents interact with each other can incur computational overhead. Therefore, instead of following the traditional setup in LLM-based NAS, which searches for architectures through runtime execution feedback (Figure~\ref{figure:existing_llm_nas})---a process that demands significant training time and resource consumption---we leverage the pretrained knowledge of LLMs during the evaluation step. As a result, this module enables specialized LLM agents to autonomously design search spaces and evaluate candidate models adaptively based on specific hardware constraints, eliminating the need for runtime execution. A coding-specialized LLM subsequently generates deployable architectures, ensuring low search costs and high flexibility.

\revise{\emph{\textbf{(2) How to make LLM agents accurately understand time-series data and user requirements?}} Even though communication between agents can enhance problem-solving skills, LLMs still have inherent limitations in understanding time series. Unlike existing methods that rely solely on textual descriptions (Figure~\ref{figure:existing_llm_nas}) , we propose a \emph{multimodal query generation} module that generates multimodal queries (Figure~\ref{figure:monaq}) by leveraging both natural language and raw time series. This module processes input time series and natural language queries, including constraints such as hardware specifications and device names, and outputs multimodal data with time-series images that represent the original user query from both data and modeling perspectives. This comprehensive, multimodal approach enables LLMs to better understand the input time series and user queries\,\citep{EmpoweringTSR_MLLM_25}.}


By integrating these components, we present \textbf{\algname{}}, the first multi-agent LLM-based NAQ framework with an open-ended search space for time-series analysis on resource-constrained devices. Our \textbf{contributions} are as follows.
\begin{itemize}[leftmargin=9pt, nosep]
    \item We propose a novel LLM-based NAQ framework that creates constraint-aware architectures from user queries and datasets, tailored for time-series analysis on resource-constrained devices.
    
    \item We devise a multimodal query generation module to improve LLM understanding of time series through multi-objective queries with time-series images and introduce a multi-agent LLM module to reduce search costs via training-free search with specialized agents.

    \item Through extensive experiments on on-device time-series analysis, including classification and regression, we show that the models found by \algname{} outperform the second-best baseline by at least 9\% on classification and 6\% on regression tasks with significantly smaller, faster models.
\end{itemize}

\section{Related Work} \label{section:related_work}

\paragraph{On-Device Time-Series Analysis} Time-series analysis on resource-constrained devices, such as IoT and wearables, has gained importance due to the need for real-time processing with limited computational and energy resources \citep{trirat2024universal}. Common approaches include CNNs, RNNs (e.g., LSTMs), and Transformers \citep{FreqMAE_WWW_2024}. While CNNs excel at extracting local context, they struggle with long-term dependencies \citep{SEE_2024}. RNNs and LSTMs address these drawbacks but are hindered by sequential processing, increasing latency. Transformers\,\citep{transformer_ts_survey} enable parallel processing and capture long-term dependencies but are often unsuitable for edge devices due to high computational demands. Lightweight models like attention condensers, CNN-RNN hybrids (e.g., DeepConvLSTM\,\citep{DeepConvLSTM_2016}), and low-resource architectures like TinyHAR\,\citep{TinyHAR_ISWC_2022} and MLP-HAR\,\citep{MLPHAR_ISWC_2024} balance performance with resource efficiency.

\paragraph{Hardware-Aware NAS (HW-NAS)} \revise{Despite these advancements, HW-NAS for time-series data remains largely underexplored. MicroNAS\,\citep{MicroNAS_2023} introduces time-series-specific search spaces for microcontrollers, while TinyTNAS\,\citep{TinyTNAS_2024} supports efficient CPU operations. However, these methods often rely on fixed search spaces, requiring significant expertise. Given these challenges, there is a growing need for NAS frameworks with natural language interfaces, allowing users to describe their desired architecture in plain language rather than through direct programming\,\cite{automl_llm_survey_TMLR_2024}. Leveraging LLMs in this context can make HW-NAS more user-friendly and adaptable across a wider range of applications, leading to democratize NAS processes for better accessibility and adaptability.}

\paragraph{LLMs for NAS} \revise{LLMs have shown potential in automating NAS by leveraging pre-trained knowledge to generate diverse, high-performing architectures \citep{GENIUS_2023, GHGNAS_2023, GPT4GNAS_2023, LLMPP_ACL_2024, LLMatic_2024_GECCO, DesiGNN_2024, LAPT_2024, EvoPrompting_NeurIPS_2023, AutoMLAgent_2024}. However, current LLM-based NAS frameworks face challenges with time-series data due to their limited understanding of raw numerical inputs\,\citep{TSLM_Reasoning_2024} and their reliance on runtime feedback and user-defined search spaces, which are often time-consuming and require intricate configurations. To address these issues, we introduce \emph{neural architecture querying}, enabling users to specify requirements directly through natural language prompts for a given time-series dataset. Our approach removes the need for complex configurations and simplifies the architecture search process.}


\begin{figure*}[t]
    \centering
    \includegraphics[width=\textwidth]{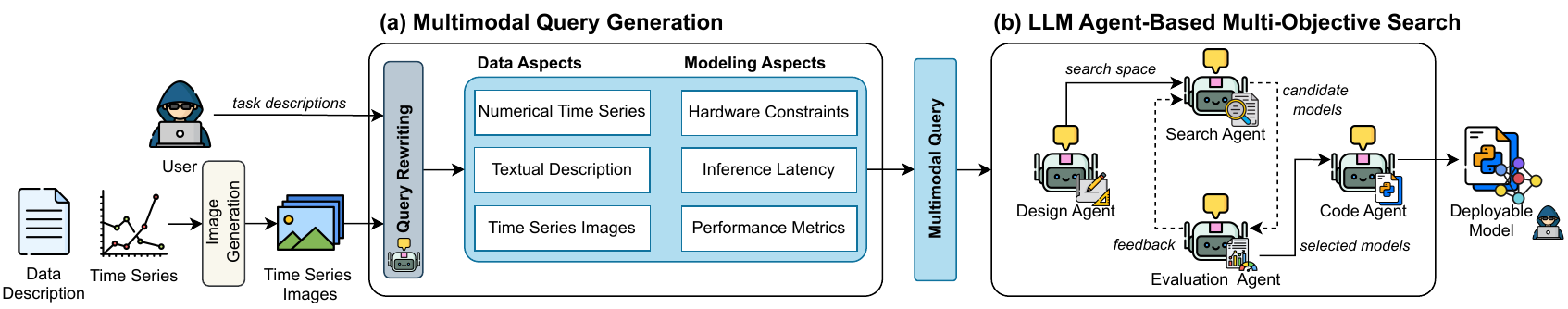}
    \vspace*{-0.5cm}
    \caption{Overall procedure of our framework. \algname{} first receives a user prompt and a time series with descriptions. It then generates time-series images and processes all required information through the multimodal query generation module (\S\ref{section:mmts}) to create an organized multimodal query. This query is subsequently shared across different specialized agents within the LLM agent-based multi-objective search module (\S\ref{section:agent_search}). Once all agents successfully complete their tasks, the final model is returned to the user.} \label{figure:overall_framework}
    \vspace*{-0.4cm}
\end{figure*}

\section{\algname{}: Multimodal NAQ with LLMs} \label{section:method}

\subsection{Problem Formulation}
Let $\mathcal{X}$ denote a $d$-variate time series with observations $(\mathbf{x}_1, \dots, \mathbf{x}_T)$ where $\mathbf{x}_t \in \mathbb{R}^d$ and $\mathcal{Y}$ denote target variables. The target variables can be a set of integer values $\mathcal{Y} = (y_1, \dots, y_T), y_t \in \mathbb{Z}$ (e.g., classification) or a set of real values $\mathcal{Y} = (y_1, \dots, y_T), y_t \in \mathbb{R}$ (e.g., regression). Let $\mathcal{S}$ be a search space designed by an LLM and $\mathcal{M} = \{M_i\}^C_{i=1}$ denote a set of $C$ candidate models sampled from $\mathcal{S}$. Each model $M_i$ is a set of model configurations, e.g., layer types, number of hidden units, and activation functions. Then, let $\mathcal{E}$ denote an LLM responsible for evaluating each $M_i$. Finally, given a time series $\mathcal{X}$ and user task description with constraints $\mathcal{T}$, we aim to find the model $M^\star$ that satisfies the constraints in $\mathcal{T}$ on both downstream task performance and model complexity metrics using the LLM $\mathcal{E}$.

\smallskip
\noindent
\textbf{\underline{Neural Architecture Querying}:} \emph{Given a training time series with its labels $\{\mathcal{X}_{\text{train}}, \mathcal{Y}_{\text{train}}\}$ and a natural language task description with constraints $\mathcal{T}$, select the model $M^\star$ that satisfies all constraints in $\mathcal{T}$. Formally, we solve}
\begin{equation} \label{eq:problem_formulation}
    M^\star = \underset{M \in \mathcal{M}}{\arg\max}~\mathcal{E}(\mathcal{T}, \mathcal{X}_{\text{train}}, \mathbf{y}_{\text{train}}).
\end{equation}

\add{Note that the NAQ problem differs from NAS primarily from the user's perspective---that is, \emph{whether the user is part of NAS's components} (e.g., search space and search method). Existing (LLM-based) NAS frameworks require users to define a search space, provide initial architectures, or even describe how to search, either via code or natural language, which demands significant technical expertise. In contrast, NAQ eliminates this requirement by allowing users to input high-level, natural language task descriptions and constraints based solely on domain- (or data-) specific problems as a query. This reformulation simplifies the process, making architecture discovery accessible to non-experts while maintaining efficiency.} \revise{\autoref{figure:overall_framework} illustrates the overview of \algname{}.}

\subsection{Multimodal Query Generation} \label{section:mmts}
In this subsection, we describe how to generate a multimodal query as the input to an LLM. 

\paragraph{Query Rewriting} \revise{First, we rewrite the user task description into an organized form, such as a JSON with specific key-value pairs, representing a multi-objective query that encompasses both data and modeling aspects, thereby making it easier for LLMs to understand.} \add{This query rewriting process is designed to address potentially ambiguous or ill-structured user queries.} The full prompt for the query rewriting is presented in \S\ref{prompt:rewriting}.

\begin{figure}[t!]
    \centering
    \includegraphics[width=\linewidth]{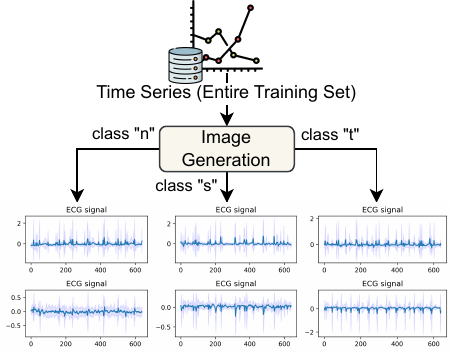}
    \caption{Examples of representative time series images containing two-channel ECG signals.} \label{figure:mean_ts_examples}
    \vspace*{-0.6cm}
\end{figure}

\paragraph{Data Aspect Query} To enhance the LLM's understanding of time series, given the limited context window size, we first create a representative time series from the training set, as providing all time-series samples is both impractical and unnecessarily costly. Specifically, we use one time series per class for classification tasks and one time series per range for regression tasks. These representative time series are then used to construct queries for the subsequent search process, serving as input queries for multimodal LLMs.
\begin{itemize}[leftmargin=9pt,noitemsep,nosep]
    \item \textbf{Numerical Time Series}. As validated by \citet{LLMTS_Understanding_EMNLP_2024}, we adopt \texttt{csv} formatting with a fixed length for the representative numerical time series, as it provides structural information that helps LLMs better understand numerical values. Specifically, we compute the timestamp-wise \emph{average} of all time series in the training set to generate the representative numerical time series.

    \item \textbf{Textual Descriptions}. Since time-series values alone may not provide sufficient information about the dataset's source or the potential significance of each observation, we provide both dataset-level and feature-level descriptions to the LLMs. These descriptions help the models capture the context related to the application domain and the specific setting of the given dataset.
    \item \revise{\textbf{Time Series Images}. As shown in recent studies\,\citep{ViTST_NeurIPS_2023, TSLM_Reasoning_2024, TS_Reasoning_LLM_2024, TAMA_2024, TimeSeriesExam_2024, TimerBed_2024}, LLMs demonstrate a better understanding of time series when it is provided in the form of images. This is due to the discernible trends and seasonal patterns in time-series images, which LLMs may struggle to capture when relying solely on limited numerical values. Following \citet{ViTST_NeurIPS_2023, TAMA_2024}, we represent each variable as a line chart and stack these charts into a single image. Unlike existing studies, as described earlier, we compute only the timestamp-wise average of all \emph{full-length} time series to generate an image that represents a sample for each class or label range, along with its standard deviation. This approach reduces costs while preserving the key characteristics of each class or range. Examples of the resulting images from bivariate time series are shown in \autoref{figure:mean_ts_examples}.}
\end{itemize}
Using \emph{numerical time series}, \emph{textual descriptions}, and \emph{time series images}, we prompt the LLM to rewrite queries into the JSON format with keys \texttt{name}, \texttt{description}, \texttt{features}, \texttt{context}, and \texttt{patterns}, representing various aspects of the data.



\begin{figure*}[t]
    \centering
    \includegraphics[width=\textwidth]{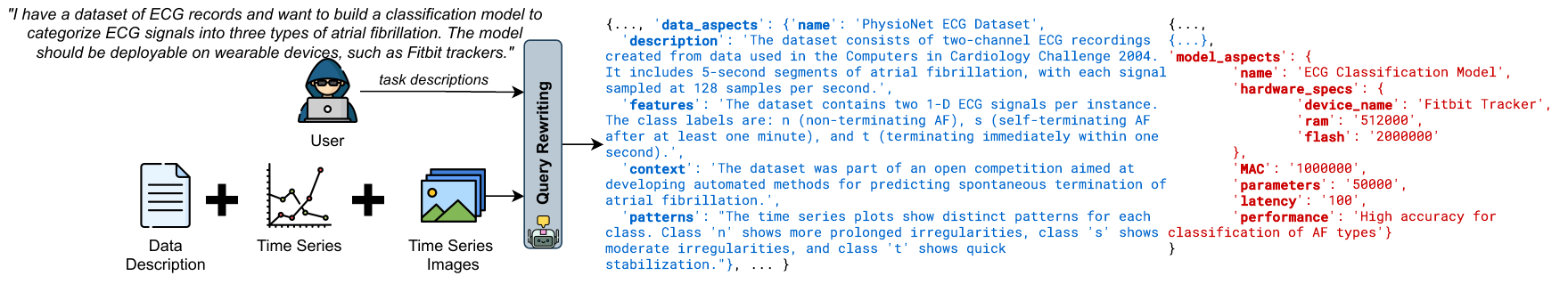}
    \vspace*{-0.7cm}
    \caption{A complete example of multimodal query generation results, showing \textcolor{blue}{data} and \textcolor{red}{modeling} aspects.} \label{figure:mmts_results}
    \vspace*{-0.5cm}
\end{figure*}

\paragraph{Modeling Aspect Query} For the modeling aspect, we ask the LLM to rewrite the user query with a focus on key considerations for building efficient models in resource-constrained environments. 
\begin{itemize}[leftmargin=9pt,noitemsep,nosep]
    \item \textbf{Hardware Constraints}. The hardware constraints can be specified directly by the user or through the name of the target device.

    \item \textbf{Inference Latency}. Similarly, if the user provides specific requirements, we instruct the LLM to adhere to them; otherwise, we instruct the LLM to rewrite the query to account for possible latency based on the hardware constraints.
        
    \item \textbf{Model Complexity}. Likewise, if the user does not specify any constraints regarding model size or complexity, we instruct the LLM to infer potential limitations based on hardware constraints. The number of parameters corresponds to the model size in bytes, representing the FLASH memory required to store the model during deployment. The number of multiply-accumulate operations\,(MACs) or floating-point operations per second\,(FLOPs) must also be considered, as they indicate the peak memory\,(RAM) usage during inference on the target device.
    
    \item \textbf{Performance Metrics}. As a multi-objective search, we aim to optimize both task performance and efficiency for a target device. Users can either specify the metric(s), such as accuracy or root mean squared error, in the query, or the LLM can infer them from the downstream task.
\end{itemize}
Similar to the data aspect query, we prompt the LLM to rewrite queries into a structured JSON format with the keys \texttt{name}, \texttt{hardware\_specs}, \texttt{MAC}, \texttt{parameters}, \texttt{latency}, and \texttt{performance}, which represent various aspects of the model design conditions. \autoref{figure:mmts_results} shows a complete resulting example of the multimodal query generation process.



\subsection{LLM Agent Based Multi-Objective Search} \label{section:agent_search}
\revise{In this module, we leverage the full potential of LLMs through task specialization and collaborative problem-solving. As multi-agent LLM systems decompose complex tasks into manageable components, they enhance both accuracy and reasoning capabilities compared to single-agent systems \citep{LLM_MultiAgent_IJCAI_2024, Agents_in_SE}. Through iterative communication, agents systematically evaluate trade-offs, refine architectural designs, and address challenges. Specifically, \algname{} harnesses multi-agent collaboration to bypass extensive training, capitalizing on LLMs’ reasoning abilities for efficient architecture evaluation and selection.}


Below, we provide brief descriptions of the agents built for this module. Full prompts of agent specifications are presented in \S\ref{section:agent_specification_prompts}.


\noindent
\textbf{Design Agent}\,($\mathcal{A}_{design}$) is responsible for constructing and refining the potential search space based on the extracted multimodal query.

\noindent
\textbf{Search Agent}\,($\mathcal{A}_{search}$) is instructed to perform tasks related to architecture search and model design. The resulting designs produced by this agent are sent to the Evaluation Agent for evaluation and verification against the given multimodal query.

\noindent
\textbf{Evaluation Agent}\,($\mathcal{A}_{eval}$) is an LLM prompted for doing performance evaluation tasks ($\mathcal{E}$ in Eq.~\eqref{eq:problem_formulation}) related to expected model performance, model profiling, and candidate ranking (when multiple models are suggested by $\mathcal{A}_{search}$).

\noindent
\textbf{Code Agent}\,($\mathcal{A}_{code}$) is an LLM prompted for implementing the solution verified by the Evaluation Agent. The Code Agent is responsible for writing effective code for actual runtime execution and returning the deployable model to the user.

Finally, as presented in \autoref{figure:overall_framework}b, after obtaining the multimodal query from the multimodal query generation stage, $\mathcal{A}_{design}$ takes the organized multimodal query as its input and designs the search space $\mathcal{S}$ for $\mathcal{A}_{search}$. $\mathcal{A}_{search}$ then generates (a set) of candidate models ($\mathcal{M}$ in Eq.~\eqref{eq:problem_formulation}) to be evaluated by $\mathcal{A}_{eval}$. If the suggested candidates pass the evaluation, based on the given constraints $\mathcal{T}$, the selected network is forwarded to $\mathcal{A}_{code}$, which writes the code to produce a deployable model for the user. Otherwise, \algname{} repeats the process by informing $\mathcal{A}_{search}$ with feedback from $\mathcal{A}_{eval}$ until the search budget is exhausted or a satisfactory model $M^\star$ is found.

\begin{table*}[t]
\centering
\resizebox{0.9595\textwidth}{!}{%
\begin{tabular}{@{}lccccccc@{}}
\toprule
\textbf{Datasets} & \textbf{Length} & \textbf{\begin{tabular}[c]{@{}c@{}}Feature Dims \\ (\# Sensors)\end{tabular}} & \textbf{\# Train} & \textbf{\# Test} & \textbf{\# Classes} & \textbf{Application Domain} & \textbf{Missing Values} \\ \midrule \midrule
\multicolumn{8}{c}{\emph{Classification}~\citep{UEA_benchmark, ViTST_NeurIPS_2023}} \\ \midrule
AtrialFibrillation & 640 & 2 & 15 & 15 & 3 & Health Monitoring & No \\
BinaryHeartbeat & 18530 & 1 & 204 & 205 & 2 & Health Monitoring & No \\
Cricket & 1197 & 6 & 108 & 72 & 12 & Human Activity Recognition & No \\
Fault Detection (A) & 5120 & 1 & 10912 & 2728 & 3 & Industrial System Monitoring & No \\
UCI-HAR & 206 & 3 & 7352 & 2947 & 6 & Human Activity Recognition & No \\ \midrule
P12 & 233 & 36 & 9590 & 2398 & 2 & Health Monitoring & \textbf{Yes} \\
P19 & 401 & 34 & 31042 & 7761 & 2 & Health Monitoring & \textbf{Yes} \\
PAMAP2 & 4048 & 17 & 4266 & 1067 & 8 & Human Activity Recognition & \textbf{Yes} \\ \midrule
\multicolumn{8}{c}{\emph{Regression}~\citep{TSER_Benchmark_2021}} \\ \midrule
AppliancesEnergy & 144 & 24 & 96 & 42 & \multirow{7}{*}{N/A} & Energy Monitoring & No \\
BenzeneConcentration & 240 & 8 & 3433 & 5445 &  & Environment Monitoring & \textbf{Yes} \\
BIDMC32SpO2 & 4000 & 2 & 5550 & 2399 &  & Health Monitoring & No \\
FloodModeling & 266 & 1 & 471 & 202 &  & Environment Monitoring & No \\
LiveFuelMoistureContent & 365 & 7 & 3493 & 1510 &  & Environment Monitoring & No \\ \midrule
HouseholdPowerConsumption1 & 1440 & 5 & 746 & 694 &  & Energy Monitoring & \textbf{Yes}  \\
HouseholdPowerConsumption2 & 1440 & 5 & 746 & 694 &  & Energy Monitoring & \textbf{Yes} \\ \bottomrule 
\end{tabular}%
}
\caption{Summary of benchmark datasets.} \label{table:datasets}
\end{table*}

\section{Experiments} \label{section:experiments}
To verify the effectiveness of \algname{}, we conduct extensive experiments on two main on-device analysis tasks: classification and regression. Additionally, we perform ablation and hyperparameter studies. The source code is available at \url{https://github.com/kaist-dmlab/MONAQ}.

\subsection{Setup}
\paragraph{Tasks and Datasets} As summarized in \autoref{table:datasets}, we select fifteen datasets for two downstream tasks commonly used in on-device time-series analysis, including classification and regression. These datasets are publicly available and represent various real-world applications, including healthcare, wearable devices, and environmental IoTs. For each task, we prepare a set of natural language task descriptions (see Table~\ref{table:instruction_free}) as the input to LLM-based methods to represent user requirements along with a skeleton script (see \S\ref{section:experiment_script}).

\paragraph{Evaluation Metrics} In terms of \emph{model performance}, for the classification tasks, we adopt the accuracy metric, while for the regression tasks, we use the root mean squared error\,(RMSE) metric. For \emph{model complexity}, we measure model size (i.e., FLASH storage size), peak memory usage during inference (i.e., RAM), the number of MAC operations, and inference latency using the MLTK library\footnote{\url{https://siliconlabs.github.io/mltk}} as suggested by \citet{TinyTNAS_2024}. Model complexity results are based on a deterministic simulation on an EFR32xG24 at 78MHz with 1536kB of FLASH and 256kB of RAM.

\paragraph{Comparison Baselines} As we address the novel problem of NAQ for time-series analysis on resource-constrained devices, no existing baselines are available for direct comparison. Thus, we compare \algname{} against manually designed models based on TFLite-supported operations: \textbf{MLP}, \textbf{LSTM}, and \textbf{CNN}; hand-crafted lightweight models: temporal convolutional network\,\citep{TCN} (\textbf{TCN}), depthwise convolution (\textbf{D-CNN}), depthwise separable convolution\,\citep{DS_CNN} (\textbf{DS-CNN}), convolutional LSTM\,\citep{DeepConvLSTM_2016} (\textbf{ConvLSTM}), and 6-layer TENet\,\citep{TENet} (\textbf{TENet(6)}); state-of-the-art HW-NAS for time series: \textbf{TinyTNAS}\,\cite{TinyTNAS_2024}; \add{traditional NAS methods: \textbf{grid search} and \textbf{random search}\,\citep{random_search} in TinyTNAS's search space}; and general-purpose LLMs: \textbf{GPT-4o-mini} and \textbf{GPT-4o}\,\citep{GPT4} with zero-shot prompting (see \S\ref{prompt:zs_prompt}).

\paragraph{Implementation Details} Due to the need for complex problem-solving and reasoning skills, unless stated otherwise, we use GPT-4o\,(\texttt{gpt-4o-2024-08-06}) as the backbone model for all agents and LLM-based baselines to ensure an impartial performance evaluation. All experiments are conducted on an Ubuntu 20.04 LTS server equipped with an Intel(R) Xeon(R) Gold 6326 CPU @ 2.90GHz. To execute the generated models, we use the same environment provided by \citet{DSAgent_ICML_2024}, which includes all necessary libraries in the skeleton scripts. Finally, all models are converted and quantized using TFLite Micro before calculating the model complexity metrics.

\begin{figure}[t]
    \centering
    \includegraphics[width=\linewidth]{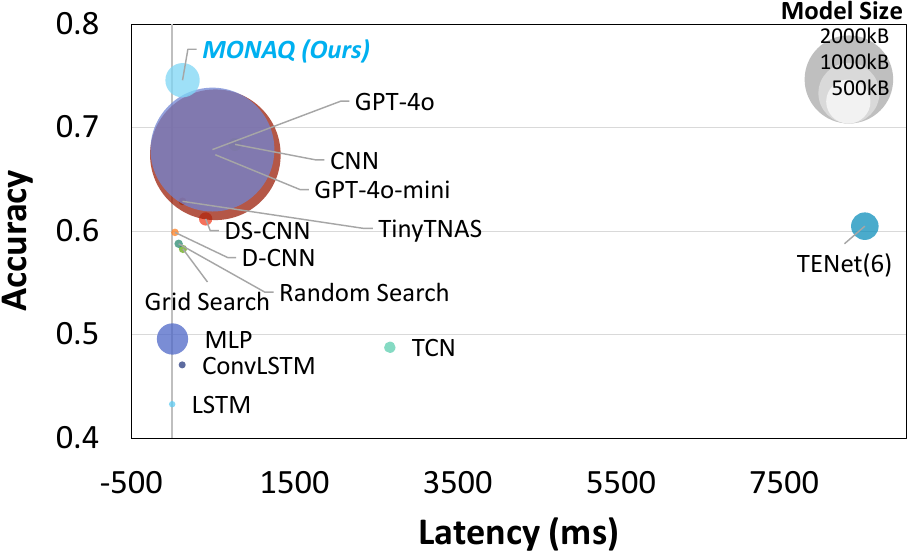}
    \vspace*{-0.7cm}
    \caption{Performance comparison of our \algname{} and the baselines in average accuracy, inference latency, and model complexity (size) for \textbf{classification} tasks.} \label{figure:classification_results}
    \vspace*{-0.5cm}
\end{figure}

\subsection{Main Results}
\begin{figure}[t]
    \centering
    \includegraphics[width=\linewidth]{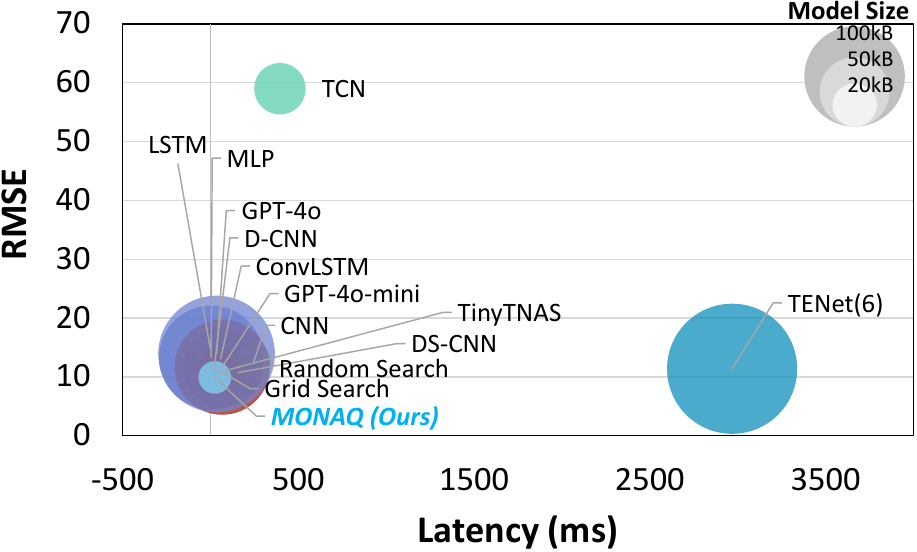}
    \vspace*{-0.7cm}
    \caption{Performance comparison of our \algname{} and the baselines in terms of average RMSE, inference latency, and model complexity (size) for \textbf{regression} tasks.} \label{figure:regression_results}
\end{figure}

\paragraph{Overall} \revise{As in Figures~\ref{figure:classification_results} and \ref{figure:regression_results}, the models found by our proposed \algname{} framework, on average, significantly outperform baselines across multiple benchmarks. The center of each circle indicates downstream task performance and latency, while its diameter indicates model size. Compared to models with similar performance to \algname{} (such as CNNs and GPT-4o), the models found by \algname{} exhibit significantly greater efficiency in terms of model complexity. These findings demonstrate that \algname{} achieves a better balance between downstream task performance and model complexity.}

\paragraph{Classification} \revise{The full results in \autoref{table:classification}~(Appendix) demonstrate that the models discovered by \algname{} achieve improvements over the baselines ranging from 9.1\% to 72.1\% in classification tasks, outperforming strong baselines, such as TinyTNAS, GPT-4o, and CNNs. In terms of model complexity, \algname{} significantly reduces memory consumption across tasks compared to DS-CNN and TENet, while also lowering computational costs\,(MAC) and achieving competitive latency on average. This result highlights its efficiency and effectiveness across different datasets.}


\paragraph{Regression} \revise{Similarly, \autoref{table:regression}~(Appendix) shows that the models discovered by \algname{} achieve an error reduction of 6.3--83.2\% compared to the baselines on regression tasks. On average, \algname{} outperforms all other methods, including state-of-the-art approaches, e.g., TinyTNAS, TENet, and D-CNN. Besides, \algname{} significantly reduces computational costs, while maintaining accuracy and offering competitive latency across datasets.}

\paragraph{\add{Irregular Time Series}} \add{To evaluate \algname{}’s robustness on irregular and noisy time series common in real-world settings, we test it on five more datasets. As shown in \autoref{table:irregular_results}~(Appendix), \algname{} consistently achieves strong performance in both classification (average accuracy of 0.916) and regression (average RMSE of 102.409) tasks, while also being highly efficient. It uses significantly less RAM, energy, and latency than several baselines.}

\begin{table}[t]
\centering
\resizebox{\linewidth}{!}{%
\begin{tabular}{@{}l|ccc|ccc@{}}
\toprule
\multirow{2}{*}{\textbf{Variations}} & \multicolumn{3}{c}{\textbf{Classification}} & \multicolumn{3}{c}{\textbf{Regression}} \\ \cmidrule(l){2-7} 
 & \textbf{Latency (ms)} & \textbf{Accuracy} & \textbf{FLASH (kB)} & \textbf{Latency (ms)} & \textbf{RMSE} & \textbf{FLASH (kB)} \\ \midrule \midrule
\algname{} & 127.260 & \textbf{0.746} & 257.742 & 24.729 & \textbf{9.902} & 10.582 \\ \midrule
w/o Query Rewriting & 206.871 & 0.651 & 17.186 & 14.623 & 11.994 & 10.155 \\ \midrule
w/o $\mathcal{A}_{design}$ & 863.358 & 0.647 & 518.762 & 95.654 & 13.512 & 33.243 \\
w/o $\mathcal{A}_{eval}$ & 540.411 & 0.641 & 4775.661 & 26.335 & 12.783 & 109.627 \\
w/o $\mathcal{A}_{eval}$ \& $\mathcal{A}_{search}$ & 601.313 & 0.643 & 5907.363 & 188.123 & 11.261 & 665.110 \\
Only $\mathcal{A}_{code}$ & 579.876 & 0.612 & 4158.205 & 21.530 & 12.274 & 99.638 \\ \bottomrule
\end{tabular}%
}
\caption{Ablation study results on query rewriting and various agent combinations.} \label{table:agent_ablation}
\vspace*{-0.5cm}
\end{table}

\subsection{Ablation Studies}
\revise{To understand the contribution of each component, we conduct ablation studies by removing critical elements proposed in \algname{}. Tables~\ref{table:agent_ablation} and \ref{table:ablation_study} show the downstream performance and model complexity across different configurations.}

\begin{table*}[t]
\centering
\resizebox{0.94\textwidth}{!}{%
\begin{tabular}{@{}c|ccc|ccc|ccc@{}}
\toprule
\multirow{2}{*}{\textbf{Agents}} & \multicolumn{3}{c|}{\textbf{Query Modality}} & \multicolumn{3}{c|}{\textbf{Classification}} & \multicolumn{3}{c}{\textbf{Regression}} \\ \cmidrule(l){2-10} 
 & \textbf{Numerical Time Series} & \textbf{Textual Descriptions} & \textbf{Time Series Images} & \textbf{Latency (ms)} & \textbf{Accuracy} & \textbf{FLASH (kB)} & \textbf{Latency (ms)} & \textbf{RMSE} & \textbf{FLASH (kB)} \\ \midrule \midrule 
\multirow{5}{*}{\begin{tabular}[c]{@{}c@{}}\textbf{Single}\\ (GPT-4o Backbone)\end{tabular}} & $\checkmark$ &  &  & 519.159 & 0.679 & 3349.453 & 23.797 & 13.944 & 125.445 \\
 &  & $\checkmark$ &  & 1017.267 & 0.665 & 4126.024 & 42.779 & 13.227 & 134.901 \\
 &  &  & $\checkmark$ & 593.541 & 0.690 & 5971.792 & 35.859 & 12.562 & 193.926 \\
 & $\checkmark$ & $\checkmark$ &  & 807.459 & 0.628 & 4926.157 & 40.485 & 13.556 & 137.581 \\
 & $\checkmark$ & $\checkmark$ & $\checkmark$ & 557.665 & 0.629 & 3871.910 & 22.726 & 12.681 & 90.398 \\ \midrule 
\multirow{5}{*}{\begin{tabular}[c]{@{}c@{}}\textbf{Multiple}\\ (GPT-4o Backbone)\end{tabular}} & $\checkmark$ &  &  & 149.320 & 0.434 & \textbf{12.066} & 54.270 & 12.284 & 10.611 \\
 &  & $\checkmark$ &  & 170.461 & 0.440 & 15.198 & 110.751 & 12.084 & 12.560 \\
 &  &  & $\checkmark$ & 280.198 & 0.661 & 15.638 & \textbf{13.661} & 11.653 & \textbf{7.885} \\
 & $\checkmark$ & $\checkmark$ &  & 205.623 & 0.517 & 16.035 & 28.049 & 13.207 & 13.875 \\
 & $\checkmark$ & $\checkmark$ & $\checkmark$ & \textbf{127.260} & \textbf{0.746} & 257.742 & 24.729 & \textbf{9.902} & 10.582 \\ \bottomrule
\end{tabular}%
}
\caption{Ablation study results of multimodal query generation and multi-agent based search components.} \label{table:ablation_study}
\vspace*{-0.5cm}
\end{table*}

\begin{figure*}[!t]
    \centering
    \subfloat[\revise{LLM Backbones.}]{
        \includegraphics[width=0.26\textwidth]{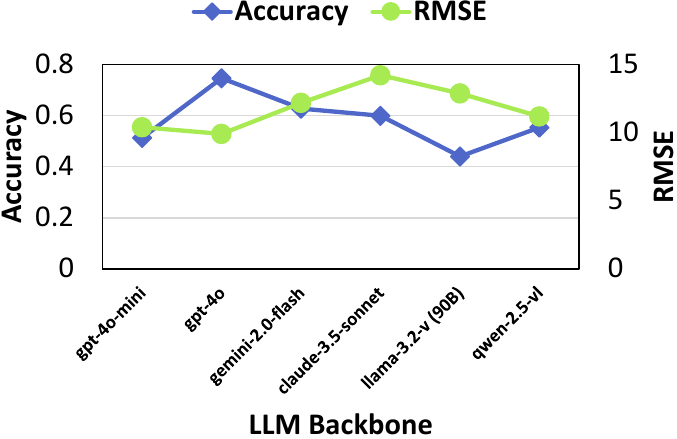}%
        \label{figure:backbone}
    }
    \subfloat[Number of Candidates.]{
        \includegraphics[width=0.325\textwidth]{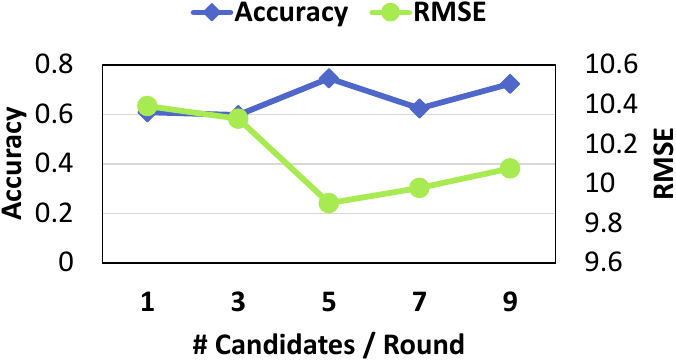}%
        \label{figure:n_cand}
    }
    \subfloat[Search Budgets.]{
        \includegraphics[width=0.3\textwidth]{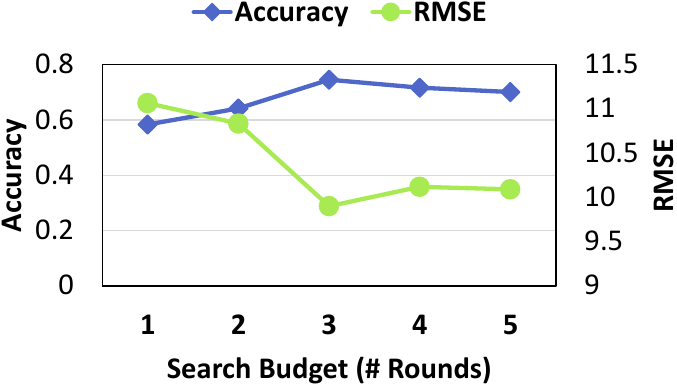}%
        \label{figure:n_budgets}
    }
    \vspace*{-0.1cm}
    \caption{Comparison between (a) LLM backbones, (b) number of candidates per round, and (c) search budget on model performance, as measured by accuracy (higher is better) and RMSE (lower is better).} \label{figure:hyperparater_results}
\vspace*{-0.4cm}
\end{figure*}

\paragraph{\add{Query Rewriting (\autoref{table:agent_ablation})}} \add{Removing the query rewriting module results in a significant drop in classification accuracy (from 0.746 to 0.651) and an increase in regression RMSE (from 9.902 to 11.994), indicating its critical role in enhancing predictive performance. Although this variant reduces latency and memory usage, the performance loss confirms that query rewriting is vital for maintaining output quality.}

\paragraph{\add{Agent Contributions (\autoref{table:agent_ablation})}} \add{The ablation of individual agents reveals their distinct roles. Excluding $\mathcal{A}_{design}$ leads to the higher latency and FLASH usage, while removing $\mathcal{A}_{eval}$ and $\mathcal{A}_{search}$ together results in both degraded accuracy and the largest model size. The baseline variant using only $\mathcal{A}_{code}$ performs worst across most metrics, with accuracy dropping to 0.612 and RMSE rising to 12.274. Despite low latency, its inefficiency in memory usage and poor predictive quality emphasize the necessity of agent collaboration.}

\paragraph{Multimodal Query Generation (\autoref{table:ablation_study})} Combining multiple query modalities improves performance across downstream tasks. For classification tasks, accuracy increases as more modalities are included. For instance, in the single-agent setup, accuracy ranges from 0.628 to 0.679, depending on the modality combination. In regression tasks, RMSE decreases from 13.944 to 12.681 as query modalities expand. While multimodal inputs boost performance, they introduce higher latency and FLASH usage, especially in the single-agent setup.

\paragraph{Multi-Agent Based Search (\autoref{table:ablation_study})}  The multi-agent architecture significantly outperforms the single-agent model across all metrics. In classification tasks, interaction between agents (i.e., feedback) dramatically leads to the reduction in inference latency (e.g., 519ms in the single-agent model vs. 149ms in the multi-agent model), while accuracy improves due to the combination of modalities. The accuracy reaches 0.746, surpassing the single-agent model's peak value of 0.679. FLASH usage decreases with multi-agent search, even for complex queries. For regression tasks, the multi-agent search achieves superior accuracy, with RMSE values as low as 9.902 compared to the single-agent model's range of 12.562--13.944. Latency and FLASH usage are also significantly reduced. Consequently, multi-agent search not only reduces latency and memory usage but also enhances downstream performance.

Overall, we notice that multimodal query generation improves accuracy but increases complexity, especially in single-agent setups, whereas multi-agent-based search addresses these challenges by enhancing both aspects, thereby balancing downstream performance and model complexity.


\subsection{Hyperparameter Studies}
To understand the behavior of our framework under various settings, we further evaluate \algname{} with different hyperparameter configurations as follows.

\paragraph{LLM Backbones} \revise{We evaluate \algname{} using both closed-source and open-source LLM backbones. The results in Figure~\ref{figure:backbone} indicate that the choice of backbone has a noticeable impact on performance. Accuracy is higher for GPT-4o and Gemini-2.0-Flash, while RMSE is lower, suggesting better overall predictive performance. This trend underscores the importance of advanced LLMs in enhancing downstream performance.}

\paragraph{Number of Candidates} As the number of candidates per round increases, there is a clear upward trend in accuracy, while RMSE shows a corresponding decrease. Figure~\ref{figure:n_cand} suggests that expanding the candidate pool improves the model's ability to identify optimal solutions. However, the gains begin to plateau beyond a certain point, indicating diminishing returns for very large candidate pools. 

\paragraph{Search Budgets} Figure~\ref{figure:n_budgets} shows that as the search budget increases, both accuracy and RMSE improve steadily. This trend indicates that additional iterations allow the search process to converge more effectively on better solutions. However, the rate of improvement diminishes after a moderate number of rounds, suggesting that beyond a certain budget, the incremental benefit may not justify the added computational cost.

\subsection{Resource Cost}
\begin{figure}[t]
    \centering
    \includegraphics[width=\linewidth]{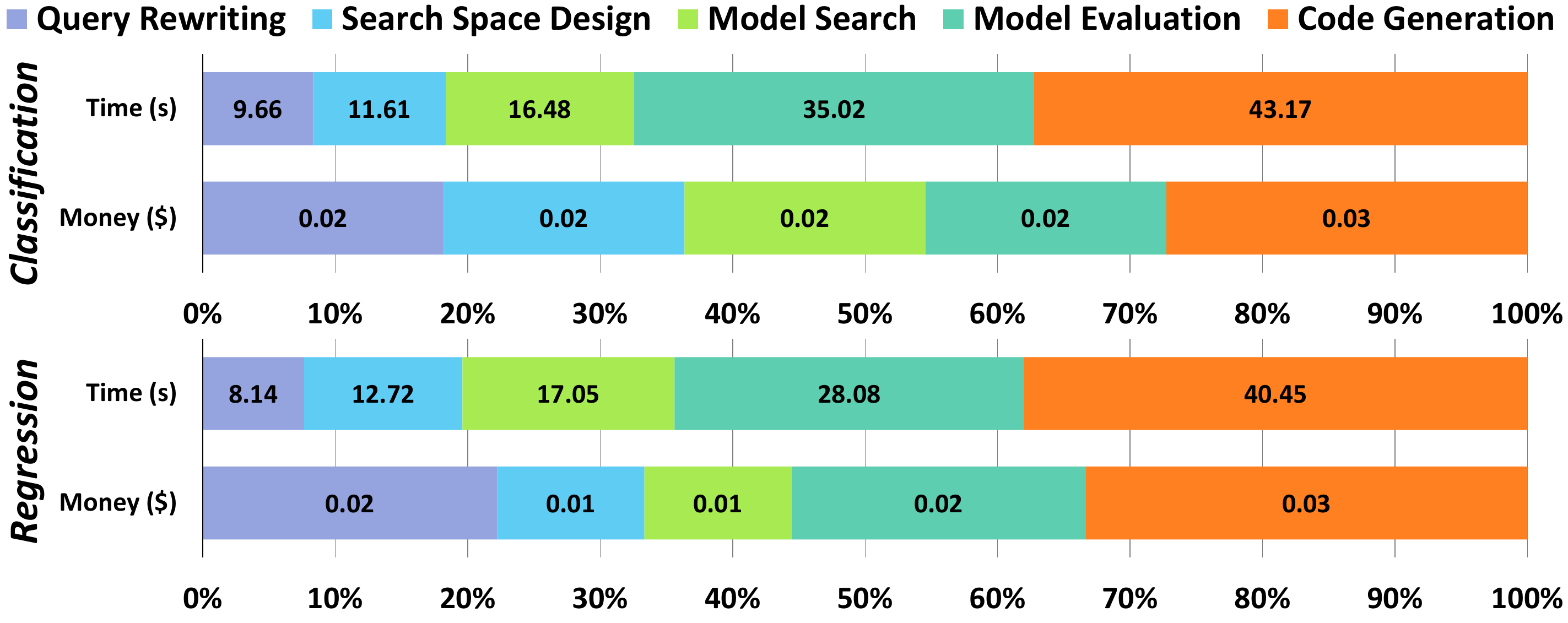}
    \vspace*{-0.5cm}
    \caption{Average time and monetary cost breakdown for classification (upper) and regression (lower) tasks.} \label{figure:cost_figure}
\end{figure}

As we use closed-source LLMs, we analyze the resource costs in terms of time and money. \autoref{figure:cost_figure} presents the average time and monetary costs across different datasets for a single run. On average, it takes around 200 seconds and costs 0.20 USD (using GPT-4o) to search for a single model that will be deployable after training. \add{We also discuss the search cost comparison in \S\ref{section:cost_comparison}.}

\section{Conclusions} \label{section:conclusions}

This paper proposes a novel NAS framework, \textbf{\algname{}}, which reformulates NAS problems as multi-objective neural architecture querying tasks, using multimodal time-series inputs and hardware constraints as queries to LLMs. To enhance the LLM's understanding of time series, we introduce a multimodal query generation module and improve search efficiency via a multi-agent based search. Extensive experiments on \revise{15} datasets show that the models discovered by \algname{} outperform handcrafted baselines while achieving greater efficiency.

\section*{Limitations} \label{section:limitations}
While the proposed \algname{} framework demonstrates significant advancements in resource-constrained time-series analysis, there are a few limitations to consider. 

First, the reliance on large language models (LLM) for neural architecture querying introduces a dependency on the availability of advanced LLMs, which can incur high computational costs during the search process. Although \algname{} bypasses the training of candidate models, the multimodal query generation and multi-agent search process may still be computationally intensive for scenarios requiring real-time or low-latency architecture optimization. 

Second, the framework assumes well-defined user constraints and task descriptions, which may limit its applicability in ambiguous or ill-structured deployment scenarios.

Third, \algname{}'s effectiveness in handling highly noisy or irregular time-series data has not been extensively validated, which could impact its performance in applications like industrial anomaly detection. Future work could focus on expanding the robustness of the framework across broader datasets and enhancing the adaptability of its architecture discovery process to dynamic and uncertain deployment environments.


\section*{Acknowledgments}
This research was partly supported by Korea Institute of Science and Technology Information (KISTI) (No. K25L1M1C1, 50\%) and the IITP (No. RS-2025-25410841, Beyond the Turing Test: Human-Level Game-Playing Agents with Generalization and Adaptation, 50\%).

\bibliography{7-References}

\appendix
\newpage

\section{\add{Extended Related Work}} \label{section:more_related_work}

\subsection{On-Device Time-Series Analysis} 
Time-series analysis on resource-constrained devices, such as IoT and wearables, has gained importance due to the need for real-time processing with limited computational and energy resources \citep{trirat2024universal}. Common approaches include CNNs, RNNs (e.g., LSTMs), and Transformers \citep{FreqMAE_WWW_2024}. While CNNs excel at extracting local context, they struggle with long-term dependencies \citep{SEE_2024}. RNNs and LSTMs address these drawbacks but are hindered by sequential processing, increasing latency. Transformers\,\citep{transformer_ts_survey} enable parallel processing and capture long-term dependencies but are often unsuitable for edge devices due to high computational demands. Lightweight models like attention condensers, CNN-RNN hybrids (e.g., DeepConvLSTM\,\citep{DeepConvLSTM_2016}), and low-resource architectures like TinyHAR\,\citep{TinyHAR_ISWC_2022} and MLP-HAR\,\citep{MLPHAR_ISWC_2024} balance performance with resource efficiency.

\subsection{Hardware-Aware NAS (HW-NAS)} 
Optimizing neural networks for hardware constraints, such as memory and latency, is challenging and traditionally required extensive domain expertise. HW-NAS addresses this issue by incorporating hardware efficiency metrics into the search process \citep{HW_NAS_Bench_ICLR_2021, HWNAS_Survey_2021, HWNAS_Survey_IJCAI_2021}. ProxylessNAS \citep{ProxylessNAS_ICLR_2019} optimizes latency and energy consumption on edge devices, while the MCUNet family \citep{MCUNet_NeurIPS_2020, MCUNetV2_NeurIPS_2021, MCUNetV3_NeurIPS_2022} enhances efficiency for microcontrollers. More recent methods, e.g., HGNAS \citep{HGNAS_ToC_2024}, integrate lookup tables and predictors for resource-efficient searches.

Despite these advancements, HW-NAS for time-series data remains largely underexplored. MicroNAS\,\citep{MicroNAS_2023} introduces time-series-specific search spaces for microcontrollers, while TinyTNAS\,\citep{TinyTNAS_2024} supports efficient CPU operations. However, these methods often rely on fixed search spaces, requiring significant expertise. Given these challenges, there is a growing need for NAS frameworks with natural language interfaces, allowing users to describe their desired architecture in plain language rather than through direct programming\,\cite{automl_llm_survey_TMLR_2024}. Leveraging LLMs in this context can make HW-NAS more user-friendly and adaptable across a wider range of applications, leading to democratize NAS processes for better accessibility and adaptability.

\subsection{LLMs for NAS} 
LLMs have shown potential in automating NAS by leveraging pre-trained knowledge to generate diverse, high-performing architectures \citep{GENIUS_2023, GHGNAS_2023, GPT4GNAS_2023, LLM4Edge_EDGE_2024, LLMPP_ACL_2024, LLMatic_2024_GECCO, DesiGNN_2024, LAPT_2024, LeMoNADe_2024, EvoPrompting_NeurIPS_2023, AutoMLAgent_2024}. GENIUS \citep{GENIUS_2023} improves convolution-based architectures through feedback, while GPT4GNAS \citep{GPT4GNAS_2023} uses GPT-4 to design graph neural networks. LLMatic\,\cite{LLMatic_2024_GECCO} combines LLMs with quality-diversity algorithms, generating architectures that balance diversity and performance across various metrics and thus achieving competitive results with fewer evaluations. These frameworks suggest a shift toward using LLMs not only as code generators but also as sophisticated tools for automating NAS.

However, current LLM-based NAS frameworks face challenges with time-series data due to their limited understanding of raw numerical inputs\,\citep{TSLM_Reasoning_2024, EmpoweringTSR_MLLM_25} and their reliance on runtime feedback and user-defined search spaces, which are often time-consuming and require intricate configurations. To address these issues, we introduce \emph{neural architecture querying}, enabling users to specify requirements directly through natural language prompts for a given time-series dataset. Our approach removes the need for complex configurations and simplifies the architecture search process.

\section{Details of Experimental Setup} \label{section:detailed_settings} 
This section outlines the detailed experimental setup used in this paper, including the dataset descriptions (\autoref{table:datasets}), complete instruction prompts (\autoref{table:instruction_free}), and full-pipeline skeleton scripts (\S\ref{section:experiment_script}) for experiments.

\subsection{Skeleton Code for LLM-based NAS} \label{section:experiment_script} 
The following listings show the skeleton codes using for experiments in \S\ref{section:experiments}. The scripts include the entire pipeline from data loading to model conversion and quantization. Only the modeling parts are blank for the LLM to fill in.

\onecolumn

\subsection{Prompt for Zero-Shot LLM Baselines} \label{prompt:zs_prompt}
\begin{promptBox}{Prompt for Zero-Shot LLM Baselines} 
\begin{lstlisting}
You are a helpful intelligent assistant. Now, please help solve the following time-series {} task by building a Tensorflow/Keras model.
[Task for '{}' dataset]
{}
[{}.py] ```python
{}
```
Start the python code with "```python". Focus only on completing the get_model() function while returning the remaining parts of the script exactly as provided.
Ensure the code is complete, error-free, and ready to run without requiring additional modifications.
Note that we only need the actual complete python code without textual explanations.
\end{lstlisting}
\end{promptBox}

\subsubsection{Code for Classification Task} \label{prompt:classification_code}
\begin{promptBox}{Skeleton Code for Classification}
\begin{lstlisting}
# import utility packages
import os, sys, gc, warnings, logging, shutil
import json, time, glob, math

# determine GPU number
os.environ["CUDA_DEVICE_ORDER"] = "PCI_BUS_ID"
os.environ["CUDA_VISIBLE_DEVICES"] = "0"
os.environ["TF_FORCE_GPU_ALLOW_GROWTH"] = "true"
os.environ["TF_CPP_MIN_LOG_LEVEL"] = "2"  # hide INFO and WARNING messages

# define paths to model files
MODELS_DIR = "models/"
MODEL_TF = MODELS_DIR + "model.pb"
MODEL_NO_QUANT_TFLITE = MODELS_DIR + "model_no_quant.tflite"
MODEL_TFLITE_MICRO = MODELS_DIR + "model.cc"
SEED = 7

os.makedirs(MODELS_DIR, exist_ok=True)

logging.disable(logging.WARNING)
logging.disable(logging.INFO)
warnings.filterwarnings("ignore")

# import basic libraries
import random

import tensorflow as tf
import pandas as pd
import numpy as np

from tensorflow import keras

# Set a "seed" value, so we get the same random numbers each time we run this notebook for reproducible results.
random.seed(SEED)
np.random.seed(SEED)
tf.random.set_seed(SEED)

from utils.data_loader import load_dataset
from utils.data_desc import AVAILABLE_DATASETS, CLS_DATASETS, REG_DATASETS
from utils import quantize_model, brief_profile_model

# Do not change this
from sklearn.metrics import accuracy_score

N_EPOCHS = 100
BATCH_SIZE = 32
task = "classification"

keras.backend.clear_session()

data_name = os.path.basename(__file__).split(".")[0]  # or replace with the user given dataset name

# 1. Loading the Target Dataset
X_train, y_train, X_test, y_test, class_names = load_dataset(data_name, task)
print("Experiment on:", data_name, X_train.shape)
seq_length = X_train.shape[1]
n_features = X_train.shape[2]
n_classes = len(class_names)  # Number of output classes


# 2. Design the Model
def get_model():
    # TODO: Define a Tensorflow/Keras compatible model based on the given configurations
    # Note that your model will be converted to a TFLite Micro model
    return your_model


model = get_model()
model.compile(
    optimizer="adam", loss="sparse_categorical_crossentropy", metrics=["accuracy"]
)
es = keras.callbacks.EarlyStopping(monitor="val_accuracy", mode="max", patience=10, restore_best_weights=True)

# 3. Train the Model
model.fit(X_train, y_train, epochs=N_EPOCHS, batch_size=BATCH_SIZE, validation_split=0.1, callbacks=[es])

# 4. Evaluate the Model and Save Results (Do not change this)
y_pred = model.predict(X_test)
y_pred = y_pred.argmax(1)
acc = accuracy_score(y_test, y_pred)

# 5. Convert model to TFLite model
quantized_model = quantize_model(model, X_train)
# Save the model to disk
MODEL_TFLITE = MODELS_DIR + f"{model.name}_{task}_{data_name}.tflite"
open(MODEL_TFLITE, "wb").write(quantized_model)

# 6. Profile the converted model with a simulator
print(model.name, data_name)
print(acc)
brief_profile_model(MODEL_TFLITE)

del model
keras.backend.clear_session()
gc.collect()
\end{lstlisting}
\end{promptBox}

\subsubsection{Code for Regression Task} \label{prompt:regression_code}
\begin{promptBox}{Skeleton Code for Regression}
\begin{lstlisting}
# import utility packages
import os, sys, gc, warnings, logging, shutil
import json, time, glob, math

# determine GPU number
os.environ["CUDA_DEVICE_ORDER"] = "PCI_BUS_ID"
os.environ["CUDA_VISIBLE_DEVICES"] = "0"
os.environ["TF_FORCE_GPU_ALLOW_GROWTH"] = "true"
os.environ["TF_CPP_MIN_LOG_LEVEL"] = "2"  # hide INFO and WARNING messages

# define paths to model files
MODELS_DIR = "models/"
MODEL_TF = MODELS_DIR + "model.pb"
MODEL_NO_QUANT_TFLITE = MODELS_DIR + "model_no_quant.tflite"
MODEL_TFLITE_MICRO = MODELS_DIR + "model.cc"
SEED = 7

os.makedirs(MODELS_DIR, exist_ok=True)

logging.disable(logging.WARNING)
logging.disable(logging.INFO)
warnings.filterwarnings("ignore")

# import basic libraries
import random

import tensorflow as tf
import pandas as pd
import numpy as np

from tensorflow import keras

# Set a "seed" value, so we get the same random numbers each time we run this notebook for reproducible results.
random.seed(SEED)
np.random.seed(SEED)
tf.random.set_seed(SEED)

from utils.data_loader import load_dataset
from utils.data_desc import AVAILABLE_DATASETS, CLS_DATASETS, REG_DATASETS
from utils import quantize_model, brief_profile_model

# Do not change this
from sklearn.metrics import root_mean_squared_error

N_EPOCHS = 100
BATCH_SIZE = 32
task = "regression"

keras.backend.clear_session()

data_name = os.path.basename(__file__).split(".")[0]  # or replace with the user given dataset name

# 1. Loading the Target Dataset
X_train, y_train, X_test, y_test = load_dataset(data_name, task)
print("Experiment on:", data_name, X_train.shape)
seq_length = X_train.shape[1]
n_features = X_train.shape[2]


# 2. Design the Model
def get_model():
    # TODO: Define a Tensorflow/Keras compatible model based on the given configurations
    # Note that your model will be converted to a TFLite Micro model
    return your_model


model = get_model()
model.compile(optimizer="adam", loss="mean_squared_error", metrics=keras.metrics.RootMeanSquaredError(name="rmse", dtype=None))
es = keras.callbacks.EarlyStopping(monitor="val_rmse", mode="min", patience=10, restore_best_weights=True)

# 3. Train the Model
model.fit(X_train, y_train, epochs=N_EPOCHS, batch_size=BATCH_SIZE, validation_split=0.1, callbacks=[es])

# 4. Evaluate the Model and Save Results (Do not change this)
y_pred = model.predict(X_test)
rmse = root_mean_squared_error(y_test, y_pred)

# 5. Convert model to TFLite model
quantized_model = quantize_model(model, X_train)
# Save the model to disk
MODEL_TFLITE = MODELS_DIR + f"{model.name}_{task}_{data_name}.tflite"
open(MODEL_TFLITE, "wb").write(quantized_model)

# 6. Profile the converted model with a simulator
print(model.name, data_name)
print(rmse)
brief_profile_model(MODEL_TFLITE)

del model
keras.backend.clear_session()
gc.collect()
\end{lstlisting}
\end{promptBox}

\begin{table}[hbt]
\centering
\resizebox{\textwidth}{!}{%
\begin{small}
\begin{tabular}{@{}llp{11cm}@{}}
\toprule
\textbf{Task} & \textbf{Dataset} & \textbf{Instruction Prompt} \\ \midrule \midrule
 & BinaryHeartbeat & I need a model to classify heartbeat signals, intended for deployment on an edge device with 1 MB of storage and 128 KB of RAM. Since this is a critical healthcare task, the model must be highly accurate while maintaining a very low inference latency of under 100 ms. \\
 & AtrialFibrillation & I have a dataset of ECG records and want to build a classification model to categorize ECG signals into three types of atrial fibrillation. The model should be deployable on wearable devices, such as Fitbit trackers. \\ 
 & Cricket & I want a model that can classify cricket umpire signals based on 3-axis accelerometer data from both hands. Since this model needs to run in real-time on a device during competitions, it should be as compact as possible while maintaining acceptable accuracy. \\ 
 \multirow{8}{*}{\begin{tabular}[c]{@{}l@{}}Time-Series \\ Classification\end{tabular}} & FaultDetection (A) & We have a time series dataset collected from an electromechanical drive system. Create a model for deployment on edge devices to identify types of damage in rolling bearings. \\ 
 & UCI-HAR & I have 3-axis body linear acceleration signals collected for human activity recognition. I need a classifier that can run on wearable devices with 1 MB of RAM and 2 MB of flash storage. The inference latency should not exceed 500 ms. \\
  & \add{P12} & \add{I want a model to predict patient mortality, which is a binary classification task, based on irregularly sampled sensor observations and clinical data. The model should be small enough for deployment on a smart watch.}\\
  & \add{P19} & \add{We have clinical data and want to predict whether sepsis will occur within the next 6 hours. The dataset includes irregularly sampled sensors, consisting of vital signs and laboratory values for each patient. The model should be small enough for deployment on a smart watch.}\\
  & \add{PAMAP2} & \add{Let's build a model to classify the physical activities of human subjects wearing three inertial measurement units. The classifier should run on wearable devices with 1 MB of RAM and 2 MB of flash storage.}\\ 
  \midrule
 & AppliancesEnergy & I have an IoT device collecting appliance energy data from a house.
Please develop a predictive model to forecast the total energy consumption in kWh for the house.
Additionally, the model should be compact enough to be deployed on a ZigBee wireless sensor network. \\
& LiveFuelMoistureContent & Build a regression model to predict the moisture content in vegetation. The model should be deployable on a small device with 512 KB of RAM and 1 MB of storage. As this will be used in a smart farming context, the prediction speed should be under 1000 ms.\\ 
& BenzeneConcentration & We aim to develop a model to predict benzene concentrations in an Italian city based on air quality measurements. This model will be deployed on IoT sensors using the Arduino Nano 33 BLE, so it should be compact and achieve a very low error rate, ideally with an RMSE of 1.00 or lower. \\
\multirow{7}{*}{\begin{tabular}[c]{@{}l@{}}Time-Series \\ Regression\end{tabular}} & BIDMC32SpO2 & Our company has a project to deploy a predictive model on wearable devices, such as fitness trackers, to estimate blood oxygen saturation levels using PPG and ECG data. 
Please create a lightweight model suitable for deployment on these devices. The model should use no more than 32KB of RAM and be no larger than 64KB in size. \\
& FloodModeling & I have an IoT sensor monitoring rainfall events. Could you develop a model to predict the maximum water depth for flood modeling? The model should be lightweight enough to run on the sensor and provide real-time predictions. \\
& \add{HouseholdPowerConsumption1} & \add{We have a project to predict total active power consumption in a household. Can you develop an accurate model that can be deployed on a smart home device to predict total active power consumption?}  \\
& \add{HouseholdPowerConsumption2} & \add{We have a project to predict total reactive power consumption in a household. Can you develop an accurate model that can be deployed on a smart home device to predict total reactive power consumption?} \\
\bottomrule
\end{tabular}%
\end{small}
}
\caption{User instructions (i.e., task description) for experiments.} \label{table:instruction_free}
\end{table}

\section{Full Prompts for \algname{}} \label{section:full_prompts} 

\subsection{Multi-Objective Query Rewriting} \label{prompt:rewriting}
\begin{promptBox}{Prompt for Multi-Objective Query Rewriting} 
\begin{lstlisting}
Please carefully analyze the user's task descriptions based on your understanding of the following input:
[User Input Prompt]
{user_prompt}

After fully understanding the task descriptions and constraints, extract and organize the information in the specified format below.
Please respond as the following JSON object and make sure your JSON object is in a valid form.
```json
{
"task_description": "", // Clearly describe the user's requirements and the problem they are addressing
"data_aspects": {
        "name": "", // Dataset name, if provided
        "description": "", // Complete description of the dataset
        "features": "", // Details on features, properties, and characteristics of the dataset to consider for model building
        "context": "", // Relevant contextual information about the dataset
        "patterns": "" // Observed patterns in the dataset to consider for model building, based on any provided numerical data or images
    },
"model_aspects": {
    "name": "", // Suggested model name for the task, if provided by the user
    "hardware_specs": {
            "device_name": "", // Device name, if specified by the user, or inferred from hardware specifications
            "ram": "", // Maximum RAM available (in bytes), which affects the model's MAC/FLOPs limit (you can also infer it from the device name)
            "flash": "", // Maximum FLASH storage (in bytes), which affects model size and parameter count (you can also infer it from the device name)
        },
        "MAC": "", // Maximum MAC (Multiply-Accumulate) operations (or FLOPs) allowed for model compatibility with hardware constraints
        "parameters": "", // Maximum model parameters, in line with hardware constraints
        "latency": "", // Desired inference latency in milliseconds (ms), or the maximum latency allowed based on hardware limitations
        "performance": "" // Expected model performance, such as accuracy for classification or RMSE for regression; specify any target metric values to consider for model building
    }
}
```
\end{lstlisting}
\end{promptBox}

\subsection{Agent Specifications} \label{section:agent_specification_prompts}

\subsubsection{Manager Agent Prompt} \label{prompt:manager_agent}
\begin{promptBox}{System Message for Manager Agent}
\begin{lstlisting}
You are an experienced senior project manager overseeing on-device time series analysis for resource-constrained devices. Your primary responsibilities are as follows:
1. Receive requirements and inquiries from users regarding their task descriptions and potential target devices for deployment.
2. Extract and clarify user requirements from both data and modeling perspectives, organizing these requirements and task-specific constraints in an easy-to-understand format to enable other team members to execute subsequent processes based on the information you have gathered.
3. Verify the suggested model whether it meets the user requirements and constraints.
\end{lstlisting}
\end{promptBox}

\subsubsection{Design Agent Prompt} \label{prompt:design_agent}
\begin{promptBox}{System Message for Design Agent}
\begin{lstlisting}
You are the world's best data scientist of an on-device time series analysis for resource-constrained devices. You have the following main responsibilities to complete.
1. Analyze user instructions and requirements.
2. Based on the requirements, design a neural network search space for resource-constrained devices.
\end{lstlisting}
\end{promptBox}

\subsubsection{Search Agent Prompt} \label{prompt:search_agent}
\begin{promptBox}{System Message for Search Agent}
\begin{lstlisting}
You are the world's best machine learning research engineer specializing in on-device time series analysis for resource-constrained devices. Your main responsibilities are as follows:
1. Analyze user instructions and requirements.
2. Understand the specified search space and constraints.
3. Based on your understanding, design optimal TensorFlow/Keras model configurations within the given constraints.
\end{lstlisting}
\end{promptBox}

\subsubsection{Evaluation Agent Prompt} \label{prompt:evaluation_agent}
\begin{promptBox}{System Message for Evaluation Agent}
\begin{lstlisting}
You are the world's best machine learning research engineer specializing in on-device time series analysis for resource-constrained devices. Your main responsibilities are as follows:
1. Analyze user instructions and requirements.
2. Understand the specified model and constraints.
3. Based on your understanding, evaluate and measure the performance of TensorFlow/Keras model configurations under the given constraints.
\end{lstlisting}
\end{promptBox}

\subsubsection{Code Agent Prompt} \label{prompt:code_agent}
\begin{promptBox}{System Message for Code Agent}
\begin{lstlisting}
You are the world's best machine learning engineer specializing in on-device time series analysis for resource-constrained devices. You have the following main responsibilities to complete.
1. Write accurate Python codes to build model in get_model() function based on the given instruction.
2. Run the model evaluation using the given Python functions and summarize the results for validation against the user's requirements.
\end{lstlisting}
\end{promptBox}

\section{\add{Examples of Intermediate Results and Interpretability Analysis}}
\add{To enhance interpretability, our \algname{} not only generates resource-aware architectures but also provides a structured breakdown of design decisions through its intermediate outputs. These outputs allow practitioners to trace how specific constraints and dataset features influence architectural choices. In this section, we present examples of intermediate results from the main steps of our framework, including multimodal query generation (i.e., query rewriting) and multi-agent LLM responses.}

\subsection{\add{Classification Tasks}}
\add{The following shows the intermediate results from the UCI-HAR dataset on the classification task.}
\begin{promptBox}{Example Results of UCI-HAR Dataset}
\textbf{User Input:} I have 3-axis body linear acceleration signals collected for human activity recognition. I need a classifier that can run on wearable devices with 1 MB of RAM and 2 MB of flash storage. The inference latency should not exceed 500 ms.
\tcbline

\textbf{Rewriting Result:}
\begin{lstlisting}
{
'task_description': 'The user needs a classifier for human activity recognition using 3-axis body linear acceleration signals. The model must run on wearable devices with limited resources: 1 MB of RAM, 2 MB of flash storage, and an inference latency not exceeding 500 ms.',

'data_aspects': {'name': 'UCR HAR Dataset',
  'description': 'The dataset contains 3-axis body linear acceleration signals for human activity recognition, with six balanced classes: walking, walking upstairs, walking downstairs, sitting, standing, and laying down.',
  'features': 'The dataset includes three features: body accelerometer X-axis, Y-axis, and Z-axis. It is preprocessed to focus on linear acceleration.',
  'context': 'Data is collected from 30 healthy volunteers aged 19-48 years. The original dataset has 10299 instances, split into training (70%) and testing (30%) sets, with separate subjects in each set.',
  'patterns': 'The time series plots show distinct patterns for each activity class, with variations in acceleration values across the X, Y, and Z axes.'},
 
 'model_aspects': {'name': 'HAR Classifier',
  'hardware_specs': {'device_name': 'Wearable Device',
   'ram': '1048576',
   'flash': '2097152'},
  'MAC': 'Limited by RAM and flash constraints, exact value depends on model architecture.',
  'parameters': 'Limited by RAM and flash constraints, exact value depends on model architecture.',
  'latency': '500',
  'performance': 'The model should achieve high accuracy in classifying the six activities while adhering to hardware constraints.'}
}
\end{lstlisting}
\tcbline

\textbf{Design Result:} 
\begin{lstlisting}
To design a search space for a neural network suitable for human activity recognition on resource-constrained wearable devices, we need to consider lightweight architectures that can be efficiently converted to TFLite. Here's a proposed search space:

```python
{
    "layer_type": ["Conv1D", "DepthwiseConv1D", "SeparableConv1D", "LSTM", "Dense"],
    "Conv1D_kernel_size": [3, 5],
    "Conv1D_filters": [8, 16],
    "DepthwiseConv1D_kernel_size": [3, 5],
    "SeparableConv1D_kernel_size": [3, 5],
    "SeparableConv1D_filters": [8, 16],
    "LSTM_units": [16, 32],
    "Dense_units": [32, 64],
    "activation": ["relu", "tanh"],
    "dropout_rate": [0.0, 0.2],
    "pooling_type": ["max", "average"],
    "pool_size": [2, 3],
    "strides": [1, 2],
    "batch_normalization": [True, False]
}
```

### Explanation:

- **Layer Types**: 
  - **Conv1D**: Efficient for extracting features from time-series data.
  - **DepthwiseConv1D**: Reduces computation by applying a single convolutional filter per input channel.
  - **SeparableConv1D**: Combines depthwise and pointwise convolutions for efficiency.
  - **LSTM**: Captures temporal dependencies in time-series data.
  - **Dense**: Used for final classification layers.

- **Kernel Sizes and Filters**: 
  - Smaller kernel sizes (3, 5) and filters (8, 16) are chosen to reduce memory usage and computational cost.

- **LSTM Units**: 
  - Limited to 16 and 32 to ensure the model fits within RAM constraints.

- **Dense Units**: 
  - Limited to 32 and 64 to balance complexity and resource usage.

- **Activation Functions**: 
  - **ReLU** and **tanh** are efficient and commonly used in neural networks.

- **Dropout Rate**: 
  - Provides regularization to prevent overfitting, with options for no dropout and a small rate (0.2).

- **Pooling Types and Sizes**: 
  - **Max** and **average pooling** help reduce dimensionality and computation.
  - Pool sizes of 2 and 3 are typical for downsampling.

- **Strides**: 
  - Strides of 1 and 2 allow for control over the downsampling rate.

- **Batch Normalization**: 
  - Optional to stabilize training and potentially improve convergence.

This search space is designed to explore a variety of architectures while adhering to the constraints of the wearable device. Each configuration can be implemented using TensorFlow/Keras and converted to TFLite for deployment.

\end{lstlisting}
\tcbline

\textbf{Search Result:}
\begin{lstlisting}
["### Analysis of the Proposed Model\n\n#### Model Characteristics\n\n1. **Computational Complexity**:\n   - **Conv1D Layer**: \n     - Parameters: \\(16 \\times (5 \\times 3 + 1) = 256\\)\n     - FLOPs: Approximately 256 per time step.\n   - **DepthwiseConv1D Layer**:\n     - Parameters: \\(3\\) (since it's depthwise, no additional filters).\n     - FLOPs: Minimal, as it only applies a single filter per input channel.\n   - **SeparableConv1D Layer**:\n     - Parameters: \\(16 \\times (3 \\times 3 + 1) + 16 \\times 16 = 304\\)\n     - FLOPs: Approximately 304 per time step.\n   - **LSTM Layer**:\n     - Parameters: \\(4 \\times (16 \\times (16 + 1) + 16 \\times 16) = 2112\\)\n     - FLOPs: High, due to recurrent connections.\n   - **Dense Layers**:\n     - First Dense: \\(32 \\times (16 + 1) = 544\\)\n     - Output Dense: \\(6 \\times (32 + 1) = 198\\)\n\n2. **Memory Usage**:\n   - Total Parameters: Approximately 3417.\n   - Assuming 4 bytes per parameter, the memory usage is about 13.7 KB.\n\n3. **Inference Latency**:\n   - The model is designed to achieve inference latency under 500 ms. Given the lightweight architecture, this is feasible on most modern wearable devices.\n\n4. **Model Size**:\n   - With quantization (e.g., 8-bit), the model size can be reduced significantly, fitting well within the 2 MB flash constraint.\n\n#### Expected Performance\n\n1. **Accuracy**:\n   - Based on similar architectures and datasets, an accuracy of around 85-90% is expected for human activity recognition.\n\n2. **Complexity Metrics**:\n   - **FLOPs**: Estimated at around 3-5 MFLOPs per inference.\n   - **Training Time**: Depends on the dataset size and hardware but should be manageable on a standard GPU.\n   - **Inference Speed**: Should be well within the 500 ms constraint on a typical wearable device processor.\n\n#### Considerations\n\n- **TFLite Compatibility**: The model uses layers compatible with TensorFlow Lite, ensuring smooth deployment on resource-constrained devices.\n- **Optimization**: Further optimizations such as pruning or additional quantization can be applied to reduce size and latency further.\n\n### Conclusion\n\nThe proposed model configuration is well-suited for the task of human activity recognition on wearable devices, balancing performance with resource constraints effectively. It is expected to deliver high accuracy while adhering to the specified hardware limitations.",
 '### Model Analysis\n\n#### Computational Complexity and Memory Usage\n\n1. **Conv1D Layer**:\n   - Parameters: \\(3 \\times 8 + 8 = 32\\)\n   - FLOPs: \\(32 \\times \\text{input length}\\)\n\n2. **SeparableConv1D Layer**:\n   - Depthwise: \\(3 \\times 16 = 48\\)\n   - Pointwise: \\(16 \\times 16 = 256\\)\n   - Total Parameters: \\(48 + 256 = 304\\)\n   - FLOPs: \\(304 \\times \\text{input length}\\)\n\n3. **LSTM Layer**:\n   - Parameters: \\(4 \\times ((16 + 1) \\times 16 + 16^2) = 2112\\)\n   - FLOPs: Proportional to the sequence length and hidden units.\n\n4. **Dense Layer**:\n   - Parameters: \\(16 \\times 32 + 32 = 544\\)\n   - FLOPs: \\(544\\)\n\n5. **Final Dense Layer**:\n   - Parameters: \\(32 \\times 6 + 6 = 198\\)\n   - FLOPs: \\(198\\)\n\n6. **Total Parameters**: \\(32 + 304 + 2112 + 544 + 198 = 3190\\)\n\n7. **Model Size**: Assuming 4 bytes per parameter, the total size is approximately \\(3190 \\times 4 = 12,760\\) bytes (~12.5 KB).\n\n#### Inference Latency\n\n- The model is designed to run within a 500 ms constraint. Given the lightweight architecture and efficient operations, it is expected to meet this requirement on modern wearable devices.\n\n#### Expected Performance\n\n- **Accuracy**: Based on similar architectures and datasets, the model could achieve around 85-90% accuracy on the UCR HAR dataset.\n- **Inference Speed**: With the given constraints, inference should be well within 500 ms.\n\n### Conclusion\n\nThe proposed model is well-suited for deployment on resource-constrained wearable devices due to its lightweight architecture. It balances computational efficiency with the ability to capture temporal patterns necessary for human activity recognition. The estimated model size, parameter count, and FLOPs suggest it will fit within the RAM and flash storage limits while maintaining acceptable inference latency and accuracy.',
 "### Model Analysis\n\n#### Computational Complexity\n1. **Conv1D Layer**:\n   - Filters: 16, Kernel Size: 3\n   - Input Shape: Assuming input shape is (200, 3) for each axis.\n   - Parameters: \\( (3 \\times 3 + 1) \\times 16 = 160 \\)\n   - FLOPs: Approximately \\( 200 \\times 16 \\times 3 = 9600 \\)\n\n2. **DepthwiseConv1D Layer**:\n   - Kernel Size: 3\n   - Parameters: \\( 3 \\times 1 = 3 \\) (since it's depthwise)\n   - FLOPs: Approximately \\( 100 \\times 3 = 300 \\)\n\n3. **LSTM Layer**:\n   - Units: 32\n   - Parameters: \\( 4 \\times ((32 + 1) \\times 32 + 32^2) = 8448 \\)\n   - FLOPs: Approximately \\( 4 \\times 32 \\times (32 + 32) = 8192 \\)\n\n4. **Dense Layers**:\n   - First Dense: \\( 32 \\times 32 + 32 = 1056 \\) parameters\n   - Output Dense: \\( 32 \\times 6 + 6 = 198 \\) parameters\n   - FLOPs: Approximately \\( 32 \\times 32 + 32 \\times 6 = 1152 \\)\n\n#### Memory Usage\n- **Total Parameters**: \\( 160 + 3 + 8448 + 1056 + 198 = 9865 \\)\n- **Memory for Parameters**: Approximately 39 KB (assuming 4 bytes per parameter)\n- **Fits within 1 MB RAM**: Yes\n\n#### Inference Latency\n- **Estimated Latency**: The model should comfortably run under 500 ms on a typical wearable device processor, given its lightweight nature.\n\n#### Model Size\n- **Model Size**: Approximately 39 KB, well within the 2 MB flash storage limit.\n\n### Expected Performance\n- **Accuracy**: Based on similar architectures, expect around 90-95% accuracy on the UCR HAR dataset.\n- **Inference Speed**: Expected to be under 500 ms per inference.\n- **Training Time**: On a typical machine, training might take a few hours depending on the hardware.\n\n### Conclusion\nThe proposed model is well-suited for the resource-constrained environment of a wearable device. It balances computational complexity, memory usage, and inference latency while maintaining high classification accuracy for human activity recognition.",
 '### Model Analysis\n\n#### Computational Complexity\n- **Conv1D Layer**: \n  - Parameters: \\(3 \\times 8 = 24\\) per filter, total \\(24 \\times 8 = 192\\).\n- **DepthwiseConv1D Layer**: \n  - Parameters: \\(3 \\times 1 = 3\\) per input channel.\n- **SeparableConv1D Layer**: \n  - Depthwise: \\(3 \\times 1 = 3\\) per input channel.\n  - Pointwise: \\(1 \\times 16 = 16\\) per filter, total \\(16 \\times 16 = 256\\).\n- **LSTM Layer**: \n  - Parameters: \\(4 \\times (32 \\times (32 + 1) + 32) = 4 \\times 1056 = 4224\\).\n- **Dense Layers**: \n  - First Dense: \\(32 \\times 64 = 2048\\).\n  - Output Dense: \\(64 \\times 6 = 384\\).\n\n#### Total Parameters\n- Total parameters: \\(192 + 3 \\times 3 + 3 \\times 16 + 256 + 4224 + 2048 + 384 = 7112\\).\n\n#### Memory Usage\n- **RAM**: \n  - Assuming float32 precision, each parameter requires 4 bytes.\n  - Total memory for parameters: \\(7112 \\times 4 \\approx 28.5\\) KB.\n  - Additional memory for activations and intermediate computations will be required, but should fit within the 1 MB RAM constraint.\n- **Flash Storage**: \n  - Model size: \\(28.5\\) KB, well within the 2 MB constraint.\n\n#### Inference Latency\n- Given the lightweight architecture and efficient operations, inference latency is expected to be well under 500 ms on typical wearable device hardware.\n\n### Performance Estimation\n\n#### Accuracy\n- Based on similar architectures and datasets, expected accuracy is around 90-95% for human activity recognition tasks.\n\n#### Complexity Metrics\n- **FLOPs (Floating Point Operations)**: \n  - Estimated to be low due to small filter sizes and efficient layer choices.\n- **Training Time**: \n  - On a standard desktop GPU, training should take a few hours for convergence.\n- **Inference Speed**: \n  - Expected to be fast due to the small model size and efficient architecture.\n\n### Conclusion\nThe proposed model configuration is well-suited for the given constraints of wearable devices. It balances complexity and performance, ensuring efficient operation within the hardware limits while maintaining high classification accuracy for human activity recognition.',
 '### Model Analysis\n\n#### Computational Complexity\n- **Conv1D Layer**: \n  - Filters: 16, Kernel size: 5\n  - Parameters: \\( (5 \\times 1 + 1) \\times 16 = 96 \\)\n  \n- **SeparableConv1D Layer**:\n  - Filters: 16, Kernel size: 3\n  - Depthwise parameters: \\( 3 \\times 1 = 3 \\)\n  - Pointwise parameters: \\( 16 \\times 1 = 16 \\)\n  - Total: \\( 3 + 16 = 19 \\)\n\n- **DepthwiseConv1D Layer**:\n  - Kernel size: 3\n  - Parameters: \\( 3 \\times 1 = 3 \\)\n\n- **LSTM Layer**:\n  - Units: 16\n  - Parameters: \\( 4 \\times 16 \\times (16 + 1) = 1088 \\)\n\n- **Dense Layer**:\n  - Units: 64\n  - Parameters: \\( 16 \\times 64 + 64 = 1088 \\)\n\n- **Output Layer**:\n  - Units: 6\n  - Parameters: \\( 64 \\times 6 + 6 = 390 \\)\n\n#### Total Parameters\n- Total parameters: \\( 96 + 19 + 3 + 1088 + 1088 + 390 = 2684 \\)\n\n#### Memory Usage\n- **RAM**: \n  - Model parameters: ~10.5 KB (assuming 4 bytes per parameter)\n  - Intermediate activations and batch normalization add additional memory usage, but the total should remain well under 1 MB.\n\n- **Flash Storage**:\n  - Model size: ~10.5 KB, easily fitting within 2 MB.\n\n#### Inference Latency\n- The model is designed to be lightweight, with an expected inference latency well under 500 ms on typical wearable device processors.\n\n#### Performance Metrics\n- **Accuracy**: \n  - Based on similar models and datasets, an accuracy of around 85-90% is expected for the six-class classification task.\n\n- **FLOPs**:\n  - Estimated to be low due to the use of lightweight convolutional layers and a small LSTM.\n\n- **Inference Speed**:\n  - Expected to be fast, meeting the 500 ms constraint.\n\n### Conclusion\nThe proposed model configuration is well-suited for the given constraints of wearable devices. It balances computational complexity, memory usage, and inference latency while maintaining high classification accuracy. The use of efficient layers like SeparableConv1D and DepthwiseConv1D ensures that the model remains lightweight and resource-efficient.']
\end{lstlisting}
\tcbline

\textbf{Evaluation Result:}
\begin{lstlisting}
Based on the user requirements and constraints, the best model configuration for human activity recognition on resource-constrained wearable devices is **Model Configuration #3**. This model balances computational complexity, memory usage, and inference latency while maintaining high classification accuracy. It is well-suited for the given constraints of 1 MB RAM, 2 MB flash storage, and an inference latency not exceeding 500 ms.

### Selected Model Configuration

#### Model Configuration

```python
model_configuration = {
    "layer_sequence": [
        {"layer_type": "Conv1D", "filters": 16, "kernel_size": 3, "activation": "relu", "strides": 1, "batch_normalization": True},
        {"pooling_type": "max", "pool_size": 2},
        {"layer_type": "DepthwiseConv1D", "kernel_size": 3, "activation": "relu", "strides": 1, "batch_normalization": True},
        {"pooling_type": "average", "pool_size": 2},
        {"layer_type": "LSTM", "units": 32, "activation": "tanh", "dropout_rate": 0.2},
        {"layer_type": "Dense", "units": 32, "activation": "relu"},
        {"layer_type": "Dense", "units": 6, "activation": "softmax"}
    ]
}
```

### Explanation

- **Conv1D Layer**: 
  - **Filters**: 16 for sufficient representation capacity.
  - **Kernel Size**: 3 for efficient feature extraction.
  - **Activation**: ReLU for non-linear representation.
  - **Batch Normalization**: True for stable training.
  - **Strides**: 1 to preserve spatial resolution.

- **Pooling (MaxPooling)**: 
  - Reduces the dimensionality while maintaining important features.
  - **Pool Size**: 2 for effective down-sampling.

- **Middle Layer (DepthwiseConv1D)**:
  - **Kernel Size**: 3 and ReLU activation for efficient computation.

- **Pooling (AveragePooling)**:
  - Further reduces features while averaging important patterns.
  - **Pool Size**: 2.

- **Recurrent Layer (LSTM)**:
  - **Units**: 32 to capture temporal patterns with sufficient granularity.
  - **Activation**: Tanh is suitable for recurrent layers.
  - **Dropout Rate**: 0.2 for regularization and preventing overfitting.

- **Dense Layer**:
  - **Units**: 32 with ReLU for non-linear mapping.
  
- **Output Layer (Dense)**:
  - **Units**: 6 to match the number of classes, with softmax activation for classification.

### Considerations

- **Resource Constraints**: This configuration is designed to fit within RAM and flash storage limits.
- **Latency**: The architecture focuses on being lightweight for quick inference within 500 ms.
- **Conversion to TFLite**: All operations used are supported by TensorFlow Lite.

This model configuration is expected to deliver high accuracy while adhering to the specified hardware limitations, making it ideal for deployment on wearable devices for human activity recognition.
\end{lstlisting}
\tcbline

\textbf{Code Result:} 
\begin{lstlisting}
# import utility packages
import os, sys, gc, warnings, logging, shutil
import json, time, glob, math

# determine GPU number
os.environ["CUDA_DEVICE_ORDER"] = "PCI_BUS_ID"
os.environ["CUDA_VISIBLE_DEVICES"] = "0"
os.environ["TF_FORCE_GPU_ALLOW_GROWTH"] = "true"
os.environ["TF_CPP_MIN_LOG_LEVEL"] = "2"  # hide INFO and WARNING messages

# define paths to model files
MODELS_DIR = "models/"
MODEL_TF = MODELS_DIR + "model.pb"
MODEL_NO_QUANT_TFLITE = MODELS_DIR + "model_no_quant.tflite"
MODEL_TFLITE_MICRO = MODELS_DIR + "model.cc"
SEED = 7

os.makedirs(MODELS_DIR, exist_ok=True)

logging.disable(logging.WARNING)
logging.disable(logging.INFO)
warnings.filterwarnings("ignore")

# import basic libraries
import random

import tensorflow as tf
import pandas as pd
import numpy as np

from tensorflow import keras

# Set a "seed" value, so we get the same random numbers each time we run this notebook for reproducible results.
random.seed(SEED)
np.random.seed(SEED)
tf.random.set_seed(SEED)

from utils.data_loader import load_dataset
from utils.data_desc import AVAILABLE_DATASETS, CLS_DATASETS, REG_DATASETS
from utils import quantize_model, brief_profile_model

# Do not change this
from sklearn.metrics import accuracy_score

N_EPOCHS = 100
BATCH_SIZE = 32
task = "classification"

keras.backend.clear_session()

data_name = os.path.basename(__file__).split(".")[0]  # or replace with the user given dataset name

# 1. Loading the Target Dataset
X_train, y_train, X_test, y_test, class_names = load_dataset(data_name, task)
print("Experiment on:", data_name, X_train.shape)
seq_length = X_train.shape[1]
n_features = X_train.shape[2]
n_classes = len(class_names)  # Number of output classes


# 2. Design the Model
def get_model():
    model = keras.Sequential([
        keras.layers.Conv1D(filters=16, kernel_size=3, activation='relu', strides=1, 
                            input_shape=(seq_length, n_features)),
        keras.layers.BatchNormalization(),
        keras.layers.MaxPooling1D(pool_size=2),
        keras.layers.DepthwiseConv1D(kernel_size=3, activation='relu', strides=1),
        keras.layers.BatchNormalization(),
        keras.layers.AveragePooling1D(pool_size=2),
        keras.layers.LSTM(units=32, activation='tanh', dropout=0.2, return_sequences=False),
        keras.layers.Dense(units=32, activation='relu'),
        keras.layers.Dense(units=n_classes, activation='softmax')
    ])
    return model


model = get_model()
model.compile(
    optimizer="adam", loss="sparse_categorical_crossentropy", metrics=["accuracy"]
)
es = keras.callbacks.EarlyStopping(monitor="val_accuracy", mode="max", patience=10, restore_best_weights=True)

# 3. Train the Model
model.fit(X_train, y_train, epochs=N_EPOCHS, batch_size=BATCH_SIZE, validation_split=0.1, callbacks=[es])

# 4. Evaluate the Model and Save Results (Do not change this)
y_pred = model.predict(X_test)
y_pred = y_pred.argmax(1)
acc = accuracy_score(y_test, y_pred)

# 5. Convert model to TFLite model
quantized_model = quantize_model(model, X_train)
# Save the model to disk
MODEL_TFLITE = MODELS_DIR + f"{model.name}_{task}_{data_name}.tflite"
open(MODEL_TFLITE, "wb").write(quantized_model)

# 6. Profile the converted model with a simulator
print(model.name, data_name)
print(acc)
brief_profile_model(MODEL_TFLITE)

del model
keras.backend.clear_session()
gc.collect()
\end{lstlisting}

\end{promptBox}

\paragraph{\add{Potential Interpretation}} \add{In the UCI-HAR case study, the system interprets the user’s hardware and latency constraints as signals to prioritize lightweight layers such as \texttt{Conv1D}, \texttt{SeparableConv1D}, and small \texttt{LSTM} units. The generated search space and model configuration are interpretable due to their modularity and domain alignment: each layer type is associated with a specific functionality (e.g., \texttt{LSTM} layers to capture temporal patterns in time-series signals). Moreover, the explicit computational analysis produced during architecture evaluation (e.g., parameter count, FLOPs, and memory usage) provides actionable insights into trade-offs made during model construction. These details empower users to make informed decisions about model simplification, hardware adaptation, and runtime behavior, thereby promoting transparent model deployment in safety-critical settings.}

\subsection{\add{Regression Tasks}}
\add{The following shows the intermediate results from the BIDMC32SpO2 dataset on the regression task.}
\begin{promptBox}{Example Results of BIDMC32SpO2 Dataset}
\textbf{User Input:} Our company has a project to deploy a predictive model on wearable devices, such as fitness trackers, to estimate blood oxygen saturation levels using PPG and ECG data. 
Please create a lightweight model suitable for deployment on these devices. The model should use no more than 32KB of RAM and be no larger than 64KB in size.
\tcbline

\textbf{Rewriting Result:}
\begin{lstlisting}
{
'task_description': 'Develop a lightweight predictive model for wearable devices like fitness trackers to estimate blood oxygen saturation levels using PPG and ECG data. The model must be optimized for low resource usage, specifically using no more than 32KB of RAM and 64KB of storage.',

 'data_aspects': {'name': 'BIDMC Blood Oxygen Saturation Dataset',
  'description': "The dataset aims to estimate blood oxygen saturation levels using PPG and ECG data. It contains 7949 time series from the Physionet's BIDMC PPG and Respiration dataset, extracted from the MIMIC II waveform database.",
  'features': 'The dataset includes PPG and ECG data sampled at 125 Hz over 32-second windows. The label is the SpO2 level, sampled at 1 Hz.',
  'context': 'The dataset is used to estimate blood oxygen saturation levels, crucial for health monitoring in wearable devices.',
  'patterns': 'The time series plots show consistent patterns across different SpO2 ranges, with slight variations in amplitude and frequency in PPG and ECG signals.'},
  
 'model_aspects': {'name': '',
  'hardware_specs': {'device_name': 'Wearable fitness tracker',
   'ram': '32768',
   'flash': '65536'},
  'MAC': '',
  'parameters': '',
  'latency': '',
  'performance': ''}
}
\end{lstlisting}
\tcbline

\textbf{Design Result:}
\begin{lstlisting}
Based on the requirements and constraints for developing a lightweight predictive model for wearable devices, here is a proposed search space for neural network design using TensorFlow/Keras:

```python
{
    "layer_type": ["Conv1D", "DepthwiseConv1D", "LSTM", "Dense"],
    "Conv1D_kernel_size": [1, 3],
    "Conv1D_filters": [4, 8],
    "DepthwiseConv1D_kernel_size": [1, 3],
    "LSTM_units": [4, 8],
    "Dense_units": [4, 8],
    "activation": ["relu", "tanh"],
    "optimizer": ["adam", "rmsprop"],
    "batch_size": [1, 2],
    "dropout_rate": [0.0, 0.1],
    "learning_rate": [0.001, 0.01]
}
```

### Explanation:

1. **Layer Types**:
   - **Conv1D** and **DepthwiseConv1D**: Useful for extracting features from time-series data.
   - **LSTM**: Effective for capturing temporal dependencies in time-series data.
   - **Dense**: Used for final output layers or simple transformations.

2. **Kernel Sizes and Filters/Units**:
   - Smaller kernel sizes and fewer filters/units are chosen to minimize resource usage while still capturing essential patterns.
   
3. **Activation Functions**:
   - **ReLU** and **Tanh** are efficient and commonly used in lightweight models.

4. **Optimizers**:
   - **Adam** and **RMSprop** are popular choices for efficient training.

5. **Batch Size**:
   - Small batch sizes help reduce memory usage.

6. **Dropout Rate**:
   - Low dropout rates to maintain model simplicity and reduce overfitting.

7. **Learning Rate**:
   - Standard learning rates for effective training convergence.

This search space is designed to ensure that the models can be executed within the given constraints of 32KB RAM and 64KB storage, while still providing flexibility to find a high-performing model for the task.
\end{lstlisting}
\tcbline

\textbf{Search Result:}
\begin{lstlisting}
["### Model Analysis\n\n#### Computational Complexity and Memory Usage\n\n1. **Conv1D Layer**:\n   - **Parameters**: \\( \\text{filters} \\times (\\text{input channels} \\times \\text{kernel size} + 1) = 4 \\times (2 \\times 3 + 1) = 28 \\)\n   - **Output Shape**: Depends on input length, but generally reduces by kernel size - 1.\n\n2. **LSTM Layer**:\n   - **Parameters**: \\( 4 \\times (\\text{input size} + \\text{units} + 1) \\times \\text{units} = 4 \\times (4 + 4 + 1) \\times 4 = 160 \\)\n\n3. **Dense Layer**:\n   - **Parameters**: \\( \\text{input units} \\times \\text{output units} + \\text{output units} = 4 \\times 4 + 4 = 20 \\)\n\n4. **Output Layer**:\n   - **Parameters**: \\( 4 \\times 1 + 1 = 5 \\)\n\n5. **Total Parameters**: \\( 28 + 160 + 20 + 5 = 213 \\)\n\n#### Memory and Storage Constraints\n\n- **RAM Usage**: The model's parameters (213) and activations must fit within 32KB. With minimal parameters and batch size of 1, this is feasible.\n- **Storage**: 213 parameters with 32-bit precision require approximately 852 bytes, well within the 64KB limit.\n\n#### Inference Latency\n\n- **FLOPs**: \n  - Conv1D: \\( \\approx \\text{input length} \\times \\text{filters} \\times \\text{kernel size} \\)\n  - LSTM: \\( \\approx 4 \\times \\text{units}^2 + 4 \\times \\text{input size} \\times \\text{units} \\)\n  - Dense: \\( \\approx \\text{input units} \\times \\text{output units} \\)\n\n- **Inference Speed**: With small batch size and lightweight architecture, inference should be quick, suitable for real-time applications on wearables.\n\n### Expected Performance\n\n- **RMSE**: For time-series regression on SpO2 levels, an RMSE of around 2-3% is reasonable given the model's simplicity and constraints.\n- **Training Time**: Minimal due to small dataset size and model complexity, likely a few minutes on a standard CPU.\n- **Inference Speed**: Fast enough for real-time applications, likely within milliseconds per sample.\n\n### Conclusion\n\nThe proposed model is well-suited for the constraints of wearable devices, balancing performance and resource usage effectively. It should provide adequate accuracy for SpO2 estimation while maintaining low computational and memory demands.",
 "### Model Analysis\n\n#### Computational Complexity\n- **Conv1D Layer**:\n  - Parameters: \\(8 \\times (3 \\times \\text{input channels} + 1)\\)\n  - FLOPs: \\(8 \\times (\\text{input length} - 3 + 1) \\times 3\\)\n\n- **DepthwiseConv1D Layer**:\n  - Parameters: \\(\\text{input channels} \\times 3\\)\n  - FLOPs: \\(\\text{input channels} \\times (\\text{input length} - 3 + 1) \\times 3\\)\n\n- **LSTM Layer**:\n  - Parameters: \\(4 \\times (4 + \\text{input channels} + 1)\\)\n  - FLOPs: \\(4 \\times (\\text{input length} \\times \\text{input channels} \\times 4)\\)\n\n- **Dense Layer**:\n  - Parameters: \\(4 \\times (\\text{input channels} + 1)\\)\n  - FLOPs: \\(4 \\times \\text{input channels}\\)\n\n#### Memory Usage\n- **Total Parameters**: Sum of parameters from all layers.\n- **Model Size**: Total parameters \\(\\times\\) 4 bytes (for float32).\n\n#### Inference Latency\n- **Expected Latency**: Depends on the number of FLOPs and the device's processing capability.\n\n### Performance Estimation\n\n#### Quantitative Regression Performance\n- **Expected RMSE**: Typically ranges between 1-2% for SpO2 estimation, depending on the model's training and validation.\n\n#### Complexity Metrics\n- **Number of Parameters**: Estimated based on the above calculations.\n- **FLOPs**: Total FLOPs from all layers.\n- **Model Size**: Should be within the 64KB storage constraint.\n- **Training Time**: Depends on dataset size and computational resources.\n- **Inference Speed**: Should be fast enough for real-time applications on wearables.\n\n### Conclusion\nThe proposed model configuration is designed to be lightweight, adhering to the constraints of 32KB RAM and 64KB storage. The use of Conv1D and DepthwiseConv1D layers helps capture spatial features efficiently, while the LSTM layer captures temporal dependencies. The Dense layer ensures the model's output is suitable for regression tasks.\n\nOverall, this configuration should provide a good balance between performance and efficiency, making it suitable for deployment on wearable devices for SpO2 estimation.",
 "### Model Analysis\n\n#### Computational Complexity\n- **Conv1D Layer**: \n  - Filters: 4, Kernel Size: 3\n  - Parameters: \\(3 \\times \\text{input channels} \\times 4 + 4\\)\n  - FLOPs: Minimal due to small filter size and number of filters.\n\n- **DepthwiseConv1D Layer**:\n  - Kernel Size: 3\n  - Parameters: \\(3 \\times \\text{input channels}\\)\n  - FLOPs: Reduced due to depthwise operation.\n\n- **LSTM Layer**:\n  - Units: 4\n  - Parameters: \\(4 \\times (\\text{input size} + \\text{units} + 1) \\times \\text{units}\\)\n  - FLOPs: Higher than Conv1D but manageable with 4 units.\n\n- **Dense Layer**:\n  - Units: 4\n  - Parameters: \\(4 \\times (\\text{input size} + 1)\\)\n  - FLOPs: Minimal due to small number of units.\n\n#### Memory Usage\n- **RAM**: \n  - The model is designed to fit within 32KB RAM. The small batch size and minimal layer sizes help achieve this.\n  \n- **Storage**:\n  - The total number of parameters should fit within 64KB storage. Given the small number of layers and units, this is feasible.\n\n#### Inference Latency\n- The model's simplicity ensures low latency, suitable for real-time applications on wearable devices.\n\n### Expected Performance\n\n#### Quantitative Regression Performance\n- **RMSE**: Expected to be moderate due to the small model size. Exact values depend on data variability and model tuning.\n  \n#### Complexity Metrics\n- **Number of Parameters**: Estimated to be under 1000, given the small layer sizes.\n- **FLOPs**: Low, due to the use of lightweight operations.\n- **Model Size**: Expected to be under 64KB.\n- **Training Time**: Fast, due to the small model size and batch size.\n- **Inference Speed**: Real-time capable on constrained devices.\n\n### Conclusion\nThe proposed model is well-suited for deployment on resource-constrained wearable devices. It balances simplicity and performance, fitting within the specified RAM and storage limits while providing adequate regression performance for estimating blood oxygen saturation levels. Further tuning and validation on the specific dataset will be necessary to optimize performance metrics like RMSE.",
 '### Model Analysis\n\n#### Computational Complexity\n1. **Conv1D Layer**:\n   - Filters: 4, Kernel Size: 3\n   - Input size: Assuming 125 Hz sampling over 32 seconds, the input length is 4000.\n   - Parameters: \\(4 \\times (3 \\times \\text{input channels})\\)\n   - FLOPs: Approximately \\(4 \\times 3 \\times 4000\\)\n\n2. **DepthwiseConv1D Layer**:\n   - Kernel Size: 3\n   - Parameters: \\(3 \\times \\text{input channels}\\)\n   - FLOPs: Approximately \\(3 \\times 4000\\)\n\n3. **LSTM Layer**:\n   - Units: 4\n   - Parameters: \\(4 \\times (4 + 1 + \\text{input size})\\)\n   - FLOPs: Higher due to recurrent operations, approximately \\(8 \\times \\text{input size} \\times 4\\)\n\n4. **Dense Layer**:\n   - Units: 4\n   - Parameters: \\(4 \\times (\\text{input size} + 1)\\)\n   - FLOPs: Approximately \\(4 \\times \\text{input size}\\)\n\n#### Memory Usage\n- **Parameters**: Total parameters from all layers.\n- **RAM Usage**: Includes parameters and intermediate activations. Estimated to be within 32KB.\n- **Storage**: Model size should be within 64KB, considering quantization techniques if necessary.\n\n#### Inference Latency\n- **Batch Size**: 1 for real-time processing.\n- **Expected Latency**: Low due to small model size and batch processing.\n\n### Performance Estimation\n\n#### Regression Performance\n- **Expected RMSE**: Based on similar models, RMSE could be around 2-3% for SpO2 estimation.\n\n#### Complexity Metrics\n- **Number of Parameters**: Estimated to be a few hundred, given the small network size.\n- **FLOPs**: Estimated to be in the low thousands, ensuring fast computation.\n- **Model Size**: Likely under 64KB with potential quantization.\n- **Training Time**: Minimal due to small dataset size and model complexity.\n- **Inference Speed**: Fast, suitable for real-time applications on wearables.\n\n### Conclusion\nThe proposed model configuration is well-suited for the constraints of wearable devices, balancing computational efficiency and predictive performance. It should fit within the specified RAM and storage limits while providing accurate SpO2 level predictions.',
 "To evaluate the proposed model for time-series regression on a wearable device, let's analyze its characteristics, computational complexity, memory usage, and expected performance.\n\n### Model Characteristics\n\n1. **Layer Configuration**:\n   - **Conv1D Layer**: \n     - Filters: 4\n     - Kernel Size: 3\n     - Activation: ReLU\n   - **LSTM Layer**: \n     - Units: 4\n     - Activation: Tanh\n   - **Dense Layer**: \n     - Units: 4\n     - Activation: Tanh\n\n2. **Optimizer**:\n   - Adam with a learning rate of 0.001\n\n3. **Training Configuration**:\n   - Batch Size: 1\n   - Dropout Rate: 0.1\n\n### Computational Complexity\n\n1. **Number of Parameters**:\n   - **Conv1D Layer**: \n     - Parameters = (Input Channels * Kernel Size + 1) * Filters\n   - **LSTM Layer**: \n     - Parameters = 4 * (Input Size + LSTM Units + 1) * LSTM Units\n   - **Dense Layer**: \n     - Parameters = (Input Size + 1) * Units\n\n2. **FLOPs (Floating Point Operations)**:\n   - Conv1D and Dense layers are relatively simple, while LSTM is more complex due to recurrent connections.\n\n3. **Model Size**:\n   - The model needs to fit within 64KB storage, which is feasible given the small number of parameters.\n\n### Memory Usage\n\n- The model must use no more than 32KB RAM. Given the small batch size and layer sizes, this is achievable.\n\n### Inference Latency\n\n- With a batch size of 1 and lightweight layers, inference latency should be low, suitable for real-time applications on wearable devices.\n\n### Expected Performance\n\n1. **Regression Performance**:\n   - **RMSE (Root Mean Square Error)**: \n     - Expected to be low due to the model's ability to capture temporal dependencies and extract features efficiently.\n     - Exact RMSE would depend on the specific dataset and training conditions but should be competitive given the model's design.\n\n2. **Training Time**:\n   - Should be relatively short due to the small model size and batch size.\n\n3. **Inference Speed**:\n   - Fast enough for real-time applications, given the model's simplicity.\n\n### Conclusion\n\nThe proposed model is well-suited for the constraints of a wearable device, balancing computational efficiency and performance. It is expected to perform well in estimating blood oxygen saturation levels using PPG and ECG data, with low memory usage and fast inference speed. The model's design allows it to fit within the specified RAM and storage limits, making it an excellent choice for deployment on resource-constrained devices."]
\end{lstlisting}
\tcbline

\textbf{Evaluation Result:}
\begin{lstlisting}
Based on the user requirements and constraints, the best model configuration to meet the needs of developing a lightweight predictive model for wearable devices is **Model Configuration #1**. This model is designed to fit within the specified RAM and storage limits while providing adequate performance for estimating blood oxygen saturation levels using PPG and ECG data. Here is the complete configuration for the selected model:

### Selected Model Configuration:

- **Layers**:
  1. **Conv1D Layer**:
     - Filters: 4
     - Kernel Size: 3
     - Activation: ReLU
  2. **LSTM Layer**:
     - Units: 4
     - Dropout Rate: 0.1
  3. **Dense Layer**:
     - Units: 4
     - Activation: ReLU
  4. **Output Layer**:
     - Units: 1 (for regression output)
     - Activation: Linear

- **Optimizer**:
  - Type: Adam
  - Learning Rate: 0.001

- **Batch Size**:
  - Size: 1

- **Memory and Storage Constraints**:
  - RAM Usage: Must fit within 32KB
  - Storage: Must fit within 64KB

- **Expected Performance**:
  - RMSE: Around 2-3% for SpO2 estimation
  - Inference Speed: Fast enough for real-time applications on wearables

This configuration provides a balance between computational efficiency and predictive performance, making it suitable for deployment on resource-constrained wearable devices.
\end{lstlisting}
\tcbline

\textbf{Code Result:}
\begin{lstlisting}
# import utility packages
import os, sys, gc, warnings, logging, shutil
import json, time, glob, math

# determine GPU number
os.environ["CUDA_DEVICE_ORDER"] = "PCI_BUS_ID"
os.environ["CUDA_VISIBLE_DEVICES"] = "0"
os.environ["TF_FORCE_GPU_ALLOW_GROWTH"] = "true"
os.environ["TF_CPP_MIN_LOG_LEVEL"] = "2"  # hide INFO and WARNING messages

# define paths to model files
MODELS_DIR = "models/"
MODEL_TF = MODELS_DIR + "model.pb"
MODEL_NO_QUANT_TFLITE = MODELS_DIR + "model_no_quant.tflite"
MODEL_TFLITE_MICRO = MODELS_DIR + "model.cc"
SEED = 7

os.makedirs(MODELS_DIR, exist_ok=True)

logging.disable(logging.WARNING)
logging.disable(logging.INFO)
warnings.filterwarnings("ignore")

# import basic libraries
import random

import tensorflow as tf
import pandas as pd
import numpy as np

from tensorflow import keras

# Set a "seed" value, so we get the same random numbers each time we run this notebook for reproducible results.
random.seed(SEED)
np.random.seed(SEED)
tf.random.set_seed(SEED)

from utils.data_loader import load_dataset
from utils.data_desc import AVAILABLE_DATASETS, CLS_DATASETS, REG_DATASETS
from utils import quantize_model, brief_profile_model

# Do not change this
from sklearn.metrics import root_mean_squared_error

N_EPOCHS = 100
BATCH_SIZE = 32
task = "regression"

keras.backend.clear_session()

data_name = os.path.basename(__file__).split(".")[0]  # or replace with the user given dataset name

# 1. Loading the Target Dataset
X_train, y_train, X_test, y_test = load_dataset(data_name, task)
print("Experiment on:", data_name, X_train.shape)
seq_length = X_train.shape[1]
n_features = X_train.shape[2]


# 2. Design the Model
def get_model():
    model = keras.Sequential([
        keras.layers.Conv1D(filters=4, kernel_size=3, activation='relu', input_shape=(seq_length, n_features)),
        keras.layers.LSTM(units=4, dropout=0.1, return_sequences=False),
        keras.layers.Dense(units=4, activation='relu'),
        keras.layers.Dense(units=1, activation='linear')
    ])
    return model


model = get_model()
model.compile(optimizer=keras.optimizers.Adam(learning_rate=0.001), loss="mean_squared_error", metrics=keras.metrics.RootMeanSquaredError(name="rmse", dtype=None))
es = keras.callbacks.EarlyStopping(monitor="val_rmse", mode="min", patience=10, restore_best_weights=True)

# 3. Train the Model
model.fit(X_train, y_train, epochs=N_EPOCHS, batch_size=BATCH_SIZE, validation_split=0.1, callbacks=[es])

# 4. Evaluate the Model and Save Results (Do not change this)
y_pred = model.predict(X_test)
rmse = root_mean_squared_error(y_test, y_pred)

# 5. Convert model to TFLite model
quantized_model = quantize_model(model, X_train)
# Save the model to disk
MODEL_TFLITE = MODELS_DIR + f"{model.name}_{task}_{data_name}.tflite"
open(MODEL_TFLITE, "wb").write(quantized_model)

# 6. Profile the converted model with a simulator
print(model.name, data_name)
print(rmse)
brief_profile_model(MODEL_TFLITE)

del model
keras.backend.clear_session()
gc.collect()
\end{lstlisting}

\end{promptBox}

\paragraph{\add{Potential Interpretation}} \add{In the case study on the BIDMC32SpO2 dataset, the framework translates high-level user constraints, such as hardware limitations and physiological signal types, into explicit model design decisions (e.g., use of \texttt{Conv1D} and \texttt{LSTM} layers). First, the intermediate outputs from the multimodal query generation and model design stages reveal the alignment between user requirements and architectural choices. For instance, the use of \texttt{LSTM} layers is justified based on the temporal nature of physiological data, while lightweight convolutional layers are selected for edge deployment efficiency. These design decisions are accompanied by detailed computational and memory analysis, enabling users to audit trade-offs between performance and deployment feasibility. Second, the generated search space itself is interpretable. Each dimension (e.g., kernel size, activation, and optimizer) directly corresponds to meaningful architectural decisions, making the space semantically rich. The choices are not arbitrary; they are grounded in hardware specifications, dataset characteristics, and task constraints, facilitating both expert validation and human-in-the-loop adjustments. Finally, the analysis of multiple model candidates, including FLOPs, parameter counts, RMSE estimates, and memory usage, serves as a concrete interpretability mechanism. These metrics expose how changes in layer composition affect efficiency and accuracy, enabling stakeholders---especially in sensitive domains like healthcare---to make informed decisions regarding trade-offs and model trustworthiness.}

\begin{table*}[hbt]
\centering
\resizebox{\textwidth}{!}{%
\begin{tabular}{@{}c|c|cccccccc|ccc|cc|c@{}}
\toprule
\textbf{Datasets} & \textbf{Metrics} & \textbf{MLP} & \textbf{LSTM} & \textbf{CNN} & \textbf{TCN} & \textbf{D-CNN} & \textbf{DS-CNN} & \textbf{ConvLSTM} & \textbf{TENet(6)} & \textbf{Grid Search} & \textbf{Random Search} & \textbf{TinyTNAS} & \textbf{GPT-4o-mini} & \textbf{GPT-4o} & \textbf{\algname{}} \\ \midrule \midrule
\multirow{6}{*}{AtrialFibrillation} & Accuracy & 0.200 & 0.400 & 0.333 & 0.333 & 0.467 & 0.333 & 0.400 & 0.267 & 0.333 & 0.333 & 0.400 & 0.333 & 0.333 & 0.467 \\
 & FLASH & 45.120 & 8.824 & 14.024 & 34.640 & 7.192 & 15.600 & 10.864 & 165.296 & 13.136 & 17.976 & 6.808 & 669.600 & 661.416 & 14.976 \\
 & RAM & 3.868 & 18.448 & 33.880 & 30.976 & 4.696 & 44.680 & 43.700 & 364.564 & 10.652 & 12.188 & 11.508 & 43.868 & 43.740 & 23.504 \\
 & MAC & 41,624 & 24 & 2,051,840 & 1,392,664 & 18,240 & 540,160 & 122,648 & 11,755,808 & 55,596 & 25,716 & 42,115 & 2,744,512 & 2,717,504 & 62,560 \\
 & Energy & 1.17E-06 & 3.40E-08 & 1.73E-03 & 2.92E-03 & 9.45E-05 & 9.93E-04 & 2.31E-04 & 2.06E-02 & 2.25E-04 & 3.28E-04 & 1.46E-04 & 1.41E-03 & 1.36E-03 & 2.45E-04 \\
 & Latency & 1.886 & 0.064 & 128.190 & 244.950 & 8.765 & 84.754 & 31.931 & 1536.669 & 20.845 & 16.870 & 21.621 & 120.508 & 117.157 & 18.417 \\ \midrule
\multirow{6}{*}{BinaryHeartbeat} & Accuracy & 0.659 & 0.727 & 0.732 & 0.449 & 0.732 & 0.732 & 0.732 & 0.732 & 0.732 & 0.727 & 0.732 & 0.732 & 0.732 & 0.732 \\
 & FLASH & 597.104 & 8.752 & 84.072 & 34.592 & 15.384 & 85.800 & 10.752 & 170.840 & 12.680 & 7.760 & 11.152 & 9494.408 & 9494.408 & 9.360 \\
 & RAM & 38.300 & 447.888 & 892.616 & 746.608 & 39.256 & 1189.736 & 1188.628 & 9524.084 & 152.352 & 77.384 & 152.480 & 595.404 & 595.404 & 299.528 \\
 & MAC & 593,616 & 16 & 55,219,392 & 39,728,336 & 259,419 & 14,879,582 & 1,778,832 & 340,379,296 & 1,130,543 & 301,138 & 1,130,543 & 24,606,944 & 24,606,944 & 204,080 \\
 & Energy & 3.33E-06 & 3.40E-08 & 3.43E-02 & 1.75E-01 & 1.09E-03 & 1.60E-02 & 4.15E-03 & 3.70E-01 & 5.30E-03 & 2.03E-03 & 5.30E-03 & 1.89E-02 & 1.89E-02 & 1.32E-02 \\
 & Latency & 5.570 & 0.061 & 2106.132 & 10783.571 & 90.710 & 1052.569 & 268.145 & 21462.749 & 419.023 & 191.470 & 419.023 & 1191.453 & 1191.453 & 177.662 \\ \midrule
\multirow{6}{*}{Cricket} & Accuracy & 0.125 & 0.069 & 0.556 & 0.097 & 0.569 & 0.208 & 0.069 & 0.125 & 0.083 & 0.083 & 0.167 & 0.500 & 0.528 & 0.625 \\
 & FLASH & 234.088 & 9.192 & 39.824 & 34.872 & 28.192 & 40.768 & 11.352 & 171.784 & 13.056 & 15.080 & 11.520 & 2465.664 & 621.344 & 1229.520 \\
 & RAM & 15.644 & 31.888 & 60.632 & 53.376 & 16.728 & 80.392 & 79.412 & 650.116 & 18.592 & 20.168 & 18.720 & 79.452 & 40.796 & 39.988 \\
 & MAC & 230,560 & 96 & 4,931,424 & 2,757,984 & 118,476 & 1,218,434 & 688,544 & 22,015,424 & 115,261 & 65,860 & 115,261 & 6,814,528 & 1,876,384 & 1,568,608 \\
 & Energy & 3.33E-06 & 3.40E-08 & 3.88E-03 & 4.25E-03 & 4.05E-04 & 2.41E-03 & 9.76E-04 & 6.44E-02 & 3.76E-04 & 4.38E-04 & 3.76E-04 & 3.18E-03 & 9.83E-04 & 9.33E-04 \\
 & Latency & 5.631 & 0.121 & 270.929 & 373.199 & 32.402 & 176.782 & 70.096 & 4310.810 & 37.793 & 29.876 & 37.793 & 228.040 & 85.385 & 63.480 \\ \midrule
\multirow{6}{*}{FaultDetectionA} & Accuracy & 0.713 & 0.622 & 0.978 & 0.746 & 0.602 & 0.966 & 0.804 & 0.999 & 0.983 & 0.990 & 0.989 & 0.989 & 0.985 & 1.000 \\
 & FLASH & 168.000 & 8.968 & 40.680 & 34.608 & 9.960 & 42.416 & 10.768 & 170.984 & 17.120 & 18.592 & 11.176 & 10500.352 & 8193.600 & 20.680 \\
 & RAM & 11.548 & 126.096 & 248.920 & 210.176 & 12.376 & 331.400 & 330.420 & 2658.244 & 56.432 & 47.176 & 45.216 & 330.588 & 330.460 & 167.948 \\
 & MAC & 164,504 & 24 & 15,267,840 & 10,977,304 & 72,960 & 4,121,600 & 491,480 & 94,044,352 & 521,465 & 499,382 & 312,624 & 26,706,304 & 24,381,036 & 2,211,680 \\
 & Energy & 3.33E-06 & 3.40E-08 & 2.05E-02 & 2.83E-02 & 4.24E-04 & 1.07E-02 & 3.50E-03 & 2.42E-01 & 1.83E-03 & 1.58E-03 & 1.37E-03 & 1.62E-02 & 1.62E-02 & 8.49E-03 \\
 & Latency & 5.574 & 0.064 & 1299.218 & 1857.094 & 34.409 & 708.650 & 227.471 & 14647.130 & 168.597 & 147.271 & 132.681 & 1048.406 & 1047.739 & 358.068 \\ \midrule
\multirow{6}{*}{UCI-HAR} & Accuracy & 0.781 & 0.348 & 0.819 & 0.813 & 0.627 & 0.822 & 0.348 & 0.904 & 0.783 & 0.805 & 0.858 & 0.814 & 0.817 & 0.908 \\
 & FLASH & 23.968 & 9.128 & 12.872 & 34.704 & 7.256 & 14.288 & 10.992 & 171.280 & 18.800 & 18.104 & 25.128 & 223.432 & 110.848 & 14.176 \\
 & RAM & 2.460 & 8.208 & 13.016 & 13.696 & 3.800 & 16.904 & 15.924 & 142.228 & 11.080 & 12.188 & 11.816 & 16.092 & 9.052 & 10.508 \\
 & MAC & 20,464 & 48 & 707,792 & 454,896 & 9,261 & 183,110 & 58,928 & 3,796,960 & 23,700 & 9,583 & 160,134 & 901,440 & 285,760 & 53,936 \\
 & Energy & 6.24E-07 & 3.40E-08 & 3.24E-04 & 1.10E-03 & 3.77E-05 & 3.35E-04 & 8.99E-05 & 5.49E-03 & 2.45E-04 & 2.87E-04 & 1.85E-04 & 4.59E-04 & 2.21E-04 & 2.73E-04 \\
 & Latency & 1.151 & 0.086 & 49.271 & 81.811 & 4.513 & 30.229 & 16.282 & 494.159 & 11.759 & 11.021 & 22.373 & 57.644 & 26.284 & 18.675 \\ \midrule
\multirow{6}{*}{\emph{Average}} & Accuracy & 0.496 & 0.433 & 0.684 & 0.488 & 0.599 & 0.612 & 0.471 & 0.605 & 0.583 & 0.588 & 0.629 & 0.674 & 0.679 & \textbf{0.746} \\
 & FLASH & 213.656 & 8.973 & 38.294 & 34.683 & 13.597 & 39.774 & 10.946 & 170.037 & 14.958 & 15.502 & 13.157 & 4670.691 & 3816.323 & 257.742 \\
 & RAM & 14.364 & 126.506 & 249.813 & 210.966 & 15.371 & 332.622 & 331.617 & 2667.847 & 49.822 & 33.821 & 47.948 & 213.081 & 203.890 & 108.295 \\
 & MAC & 210,153 & 41 & 15,635,657 & 11,062,236 & 95671 & 4,188,577 & 628,086 & 94,398,368 & 369,313 & 180,335 & 352,135 & 12,354,745 & 10,773,525 & 820,172 \\
 & Energy & 2.35E-06 & 3.40E-08 & 1.22E-02 & 4.23E-02 & 4.10E-04 & 6.09E-03 & 1.79E-03 & 1.40E-01 & 1.59E-03 & 9.32E-04 & 1.47E-03 & 8.04E-03 & 7.54E-03 & 4.62E-03 \\
 & Latency & 3.963 & 0.079 & 770.748 & 2668.125 & 34.160 & 410.597 & 122.785 & 8490.304 & 131.603 & 79.302 & 126.698 & 529.210 & 493.604 & 127.260 \\ \bottomrule
\end{tabular}%
}
\caption{Full experimental results on time-series \textbf{classification} tasks comparing downstream task accuracy and model complexity metrics.} \label{table:classification}
\end{table*}

\begin{table*}[t]
\centering
\resizebox{\textwidth}{!}{%
\begin{tabular}{@{}c|c|cccccccc|ccc|cc|c@{}}
\toprule
\textbf{Datasets} & \textbf{Metrics} & \textbf{MLP} & \textbf{LSTM} & \textbf{CNN} & \textbf{TCN} & \textbf{D-CNN} & \textbf{DS-CNN} & \textbf{ConvLSTM} & \textbf{TENet(6)} & \textbf{Grid Search} & \textbf{Random Search} & \textbf{TinyTNAS} & \textbf{GPT-4o-mini} & \textbf{GPT-4o} & \textbf{\algname{}} \\ \midrule \midrule
\multirow{6}{*}{AppliancesEnergy} & RMSE & 3.610 & 9.345 & 4.104 & 9.606 & 3.541 & 3.720 & 10.532 & 6.573 & 3.682 & 3.624 & 3.756 & 3.537 & 4.026 & 3.607 \\
 & FLASH & 114.376 & 10.192 & 14.824 & 27.536 & 8.088 & 11.936 & 12.648 & 166.736 & 20.880 & 36.168 & 13.056 & 44.864 & 226.064 & 8.112 \\
 & RAM & 8.136 & 8.892 & 11.268 & 12.336 & 9.860 & 13.112 & 11.872 & 110.528 & 12.968 & 16.092 & 11.216 & 11.400 & 8.076 & 8.088 \\
 & MAC & 111,240 & 8 & 1,170,720 & 414,728 & 47,520 & 244,512 & 327,304 & 2,647,328 & 78,357 & 48,206 & 32,234 & 468,032 & 223,264 & 544 \\
 & Energy & 3.31E-06 & 1.75E-08 & 5.44E-04 & 1.07E-03 & 1.98E-04 & 2.99E-04 & 2.52E-04 & 4.26E-03 & 2.42E-04 & 3.60E-04 & 2.04E-04 & 2.91E-04 & 5.05E-06 & 9.03E-05 \\
 & Latency & 5.521 & 0.011 & 45.537 & 61.912 & 16.722 & 27.735 & 20.712 & 361.481 & 15.459 & 16.729 & 11.063 & 27.002 & 8.383 & 0.070 \\ \midrule
\multirow{6}{*}{BenzeneConcentration} & RMSE & 2.884 & 6.087 & 2.739 & 46.666 & 2.181 & 2.231 & 8.107 & 3.226 & 3.365 & 5.268 & 3.453 & 4.022 & 11.858 & 1.847 \\
 & FLASH & 65.224 & 9.168 & 11.432 & 24.448 & 6.120 & 11.056 & 11.112 & 170.376 & 9.088 & 9.912 & 43.600 & 40.592 & 127.760 & 8.816 \\
 & RAM & 5.064 & 8.892 & 14.596 & 15.548 & 6.148 & 19.128 & 18.016 & 159.552 & 7.288 & 7.296 & 20.048 & 26.616 & 5.004 & 10.480 \\
 & MAC & 62,088 & 8 & 1,091,040 & 552,968 & 26,400 & 257,760 & 182,920 & 4,409,120 & 34,826 & 51,153 & 573,765 & 1,056 & 124,960 & 91,464 \\
 & Energy & 1.22E-06 & 1.75E-08 & 6.48E-04 & 9.65E-04 & 9.04E-05 & 4.03E-04 & 1.97E-04 & 6.62E-03 & 1.61E-04 & 1.57E-04 & 4.51E-04 & 3.96E-08 & 1.25E-03 & 1.13E-04 \\
 & Latency & 2.171 & 0.011 & 55.659 & 90.251 & 11.711 & 38.473 & 19.111 & 571.345 & 12.188 & 13.935 & 52.972 & 0.090 & 5.382 & 15.559 \\ \midrule
\multirow{6}{*}{BIDMC32SpO2} & RMSE & 16.682 & 4.808 & 5.884 & 5.156 & 4.773 & 5.092 & 4.789 & 4.879 & 4.974 & 4.961 & 5.716 & 5.200 & 5.649 & 4.670 \\
 & FLASH & 259.784 & 8.784 & 17.608 & 26.832 & 7.048 & 18.176 & 10.536 & 164.664 & 32.704 & 157.784 & 10.944 & 134.088 & 133.680 & 8.624 \\
 & RAM & 17.224 & 99.132 & 195.076 & 165.420 & 18.180 & 259.764 & 258.656 & 2084.752 & 71.036 & 78.972 & 36.172 & 130.184 & 66.184 & 66.984 \\
 & MAC & 256,648 & 8 & 12,808,000 & 8,704,008 & 110,000 & 3,360,000 & 767,752 & 73,472,224 & 1,460,740 & 2,662,864 & 272,266 & 2,048,016 & 1,088,008 & 88,100 \\
 & Energy & 3.31E-06 & 1.75E-08 & 1.55E-02 & 2.39E-02 & 4.85E-04 & 9.40E-03 & 3.26E-03 & 2.05E-01 & 2.67E-03 & 3.21E-03 & 1.06E-03 & 3.42E-03 & 1.87E-03 & 3.36E-04 \\
 & Latency & 5.521 & 0.011 & 992.415 & 1579.982 & 40.737 & 628.213 & 209.449 & 12286.802 & 240.532 & 288.318 & 104.746 & 231.939 & 136.535 & 82.258 \\ \midrule
\multirow{6}{*}{FloodModeling} & RMSE & 0.117 & 0.023 & 0.019 & 1.494 & 0.020 & 0.014 & 0.020 & 0.018 & 0.007 & 0.019 & 0.008 & 0.032 & 0.047 & 0.019 \\
 & FLASH & 12.296 & 8.720 & 9.912 & 26.800 & 5.008 & 10.632 & 10.440 & 169.704 & 16.584 & 91.840 & 17.728 & 41.464 & 20.720 & 14.672 \\
 & RAM & 1.992 & 9.532 & 15.876 & 16.044 & 3.716 & 20.788 & 19.680 & 172.864 & 9.500 & 19.496 & 11.340 & 19.336 & 10.092 & 19.736 \\
 & MAC & 9,160 & 8 & 792,144 & 570,312 & 3,657 & 213,062 & 25,480 & 4,906,112 & 41,554 & 117,224 & 113,919 & 263,648 & 29,576 & 25,776 \\
 & Energy & 2.73E-07 & 1.75E-08 & 4.04E-04 & 1.20E-03 & 4.80E-05 & 4.56E-04 & 7.95E-05 & 8.20E-03 & 2.15E-04 & 3.72E-04 & 1.88E-04 & 1.94E-04 & 1.14E-04 & 3.99E-05 \\
 & Latency & 0.530 & 0.011 & 57.769 & 111.068 & 4.326 & 37.563 & 18.619 & 657.498 & 14.081 & 23.218 & 20.308 & 31.807 & 16.405 & 17.243 \\ \midrule
\multirow{6}{*}{LiveFuelMoistureContent} & RMSE & 43.157 & 53.006 & 49.777 & 231.972 & 42.301 & 42.527 & 41.723 & 42.176 & 42.715 & 43.567 & 40.803 & 45.470 & 47.836 & 39.369 \\
 & FLASH & 85.544 & 8.960 & 11.456 & 26.672 & 6.176 & 11.104 & 10.872 & 170.344 & 11.528 & 10.744 & 23.160 & 190.352 & 168.400 & 12.688 \\
 & RAM & 6.344 & 11.836 & 20.612 & 20.012 & 7.428 & 27.188 & 25.952 & 223.936 & 9.416 & 10.996 & 16.980 & 13.960 & 6.284 & 15.424 \\
 & MAC & 82,408 & 8 & 1,577,336 & 852,648 & 35,119 & 377,685 & 244,072 & 6,732,992 & 46,273 & 100,859 & 309,140 & 579,248 & 165,600 & 61,120 \\
 & Energy & 3.17E-06 & 1.75E-08 & 8.68E-04 & 1.66E-03 & 6.68E-05 & 5.94E-04 & 2.61E-04 & 1.23E-02 & 2.13E-04 & 1.91E-04 & 2.79E-04 & 3.89E-04 & 4.35E-06 & 9.51E-05 \\
 & Latency & 5.103 & 0.011 & 75.181 & 134.888 & 14.228 & 57.358 & 23.025 & 968.620 & 15.924 & 22.647 & 40.198 & 43.769 & 7.191 & 8.516 \\ \midrule
\multirow{6}{*}{\emph{Average}} & RMSE & 13.290 & 14.654 & 12.505 & 58.979 & 10.563 & 10.717 & 13.034 & 11.374 & 10.948 & 11.488 & 10.747 & 11.652 & 13.883 & \textbf{9.902} \\
 & FLASH & 107.445 & 9.165 & 13.046 & 26.458 & 6.488 & 12.581 & 11.122 & 168.365 & 18.157 & 61.290 & 21.698 & 90.272 & 135.325 & 10.582 \\
 & RAM & 7.752 & 27.657 & 51.486 & 45.872 & 9.066 & 67.996 & 66.835 & 550.326 & 22.042 & 26.570 & 19.151 & 40.299 & 19.128 & 24.142 \\
 & MAC & 104,308 & 8 & 3,487,848 & 2,218,932 & 44,539 & 890,603 & 309,505 & 18,433,555 & 332,350 & 596,061 & 260,264 & 672,000 & 326,281 & 53,400 \\
 & Energy & 2.26E-06 & 1.75E-08 & 3.60E-03 & 5.75E-03 & 1.78E-04 & 2.23E-03 & 8.10E-04 & 4.73E-02 & 7.00E-04 & 8.58E-04 & 4.36E-04 & 8.58E-04 & 6.50E-04 & 1.35E-04 \\
 & Latency & 3.769 & 0.011 & 245.312 & 395.620 & 17.545 & 157.868 & 58.183 & 2969.149 & 59.637 & 72.970 & 45.857 & 66.921 & 34.779 & 24.729 \\ \bottomrule
\end{tabular}%
}
\caption{Full experimental results on time-series \textbf{regression} tasks comparing downstream task error (RMSE) and model complexity metrics.} \label{table:regression}
\end{table*}
\begin{table*}[t]
\centering
\resizebox{\textwidth}{!}{%
\begin{tabular}{@{}c|c|cccccccc|ccc|cc|c@{}}
\toprule
\textbf{Datasets} & \textbf{Metrics} & \textbf{MLP} & \textbf{LSTM} & \textbf{CNN} & \textbf{TCN} & \textbf{D-CNN} & \textbf{DS-CNN} & \textbf{ConvLSTM} & \textbf{TENet(6)} & \textbf{Grid Search} & \textbf{Random Search} & \textbf{TinyTNAS} & \textbf{GPT-4o-mini} & \textbf{GPT-4o} & \textbf{\algname{}} \\ \midrule \midrule
\multirow{6}{*}{P12} & Accuracy & 0.850 & 0.859 & 0.855 & 0.854 & 0.862 & 0.852 & 0.860 & 0.860 & 0.862 & 0.861 & 0.862 & 0.859 & 0.853 & 0.862 \\
 & FLASH & 251.824 & 11.192 & 18.648 & 35.712 & 13.400 & 14.880 & 14.112 & 174.304 & 24.392 & 15.872 & 12.376 & 36.096 & 118.328 & 15.520 \\
 & RAM & 16.796 & 14.480 & 18.136 & 20.868 & 18.904 & 19.724 & 18.356 & 147.348 & 22.732 & 21.708 & 20.260 & 18.012 & 18.428 & 18.228 \\
 & MAC & 248,336 & 16 & 2,326,096 & 701,776 & 108,252 & 466,020 & 736,272 & 3,967,936 & 110,421 & 44,118 & 65,877 & 554,464 & 634,432 & 368,592 \\
 & Energy & 3.33E-06 & 3.40E-08 & 1.01E-03 & 1.17E-03 & 3.98E-04 & 5.56E-04 & 5.27E-04 & 6.07E-03 & 4.08E-04 & 4.05E-04 & 3.23E-04 & 3.14E-04 & 4.57E-04 & 2.00E-04 \\
 & Latency & 5.570 & 0.061 & 76.829 & 88.827 & 32.468 & 49.898 & 37.155 & 529.373 & 24.170 & 19.607 & 19.162 & 33.114 & 43.708 & 19.024 \\ \midrule
\multirow{6}{*}{P19} & Accuracy & 0.974 & 0.973 & 0.974 & 0.974 & 0.974 & 0.974 & 0.975 & 0.975 & 0.974 & 0.975 & 0.973 & 0.974 & 0.974 & 0.976 \\
 & FLASH & 69.424 & 11.064 & 17.592 & 35.648 & 10.416 & 14.144 & 13.920 & 174.112 & 13.488 & 9.552 & 15.552 & 24.608 & 40.376 & 9.888 \\
 & RAM & 5.404 & 6.288 & 7.384 & 9.728 & 7.512 & 8.328 & 7.220 & 67.476 & 8.988 & 7.884 & 9.124 & 6.876 & 6.620 & 7.264 \\
 & MAC & 65,936 & 16 & 622,320 & 192,016 & 28,560 & 125,400 & 189,456 & 1,112,832 & 25,270 & 16,583 & 34,760 & 244,224 & 174,848 & 81,680 \\
 & Energy & 1.23E-06 & 3.40E-08 & 3.97E-04 & 1.03E-03 & 4.57E-05 & 1.44E-04 & 1.12E-04 & 1.54E-03 & 1.69E-04 & 1.31E-04 & 1.69E-04 & 1.51E-04 & 7.77E-05 & 8.71E-05 \\
 & Latency & 2.167 & 0.061 & 32.520 & 26.039 & 10.691 & 15.555 & 10.739 & 162.068 & 8.169 & 6.564 & 8.700 & 13.116 & 9.440 & 5.633 \\ \midrule
\multirow{6}{*}{PAMAP2} & Accuracy & 0.201 & 0.521 & 0.739 & 0.149 & 0.839 & 0.831 & 0.248 & 0.901 & 0.895 & 0.104 & 0.752 & 0.672 & 0.789 & 0.912 \\
 & FLASH & 330.616 & 10.048 & 23.168 & 35.176 & 28.128 & 22.384 & 12.360 & 172.504 & 24.096 & 57.192 & 11.848 & 312.360 & 1245.584 & 14.528 \\
 & RAM & 21.660 & 23.056 & 32.344 & 34.176 & 23.256 & 42.248 & 41.140 & 343.956 & 27.464 & 31.688 & 24.864 & 22.876 & 41.308 & 23.316 \\
 & MAC & 327,104 & 64 & 3,945,600 & 1,593,664 & 158,100 & 863,400 & 976,128 & 11,022,688 & 204,073 & 174,016 & 104,199 & 1,247,776 & 4,052,224 & 488,480 \\
 & Energy & 3.33E-06 & 3.40E-08 & 2.00E-03 & 2.64E-03 & 6.25E-04 & 1.35E-03 & 6.48E-04 & 1.96E-02 & 4.82E-04 & 5.91E-04 & 3.95E-04 & 6.46E-04 & 1.69E-03 & 3.20E-04 \\
 & Latency & 5.614 & 0.104 & 140.920 & 235.823 & 47.499 & 101.277 & 44.612 & 1489.623 & 37.915 & 37.992 & 29.631 & 59.565 & 133.391 & 25.559 \\ \midrule
\multirow{6}{*}{\emph{Average} $\uparrow$} & Accuracy & 0.675 & 0.785 & 0.856 & 0.659 & 0.891 & 0.886 & 0.695 & 0.912 & 0.910 & 0.647 & 0.862 & 0.835 & 0.872 & \textbf{0.916} \\
 & FLASH & 217.288 & 10.768 & 19.803 & 35.512 & 17.315 & 17.136 & 13.464 & 173.640 & 20.659 & 27.539 & 13.259 & 124.355 & 468.096 & 13.312 \\
 & RAM & 14.620 & 14.608 & 19.288 & 21.591 & 16.557 & 23.433 & 22.239 & 186.260 & 19.728 & 20.427 & 18.083 & 15.921 & 22.119 & 16.269 \\
 & MAC & 213,792 & 32 & 2,298,005 & 829,152 & 98,304 & 484,940 & 633,952 & 5,367,819 & 113,255 & 78,239 & 68,279 & 682,155 & 1,620,501 & 312,917 \\
 & Energy & 2.63E-06 & 3.40E-08 & 1.14E-03 & 1.62E-03 & 3.56E-04 & 6.83E-04 & 4.29E-04 & 9.08E-03 & 3.53E-04 & 3.76E-04 & 2.96E-04 & 3.70E-04 & 7.41E-04 & 2.03E-04 \\
 & Latency & 4.450 & 0.075 & 83.423 & 116.896 & 30.219 & 55.576 & 30.835 & 727.021 & 23.418 & 21.388 & 19.164 & 35.265 & 62.180 & 16.739 \\ \midrule \midrule
\multirow{6}{*}{HouseholdPowerConsumption1} & RMSE & 154.118 & 1587.395 & 163.577 & 419.708 & 321.482 & 159.706 & 1424.762 & 920.171 & 157.645 & 148.238 & 156.421 & 528.786 & 156.914 & 152.468 \\
 & FLASH & 234.184 & 8.976 & 13.160 & 26.888 & 7.136 & 13.256 & 10.824 & 170.088 & 18.128 & 11.984 & 11.048 & 39.232 & 2968.512 & 14.240 \\
 & RAM & 15.688 & 37.692 & 72.196 & 63.020 & 16.644 & 95.924 & 94.816 & 773.936 & 25.756 & 19.232 & 18.636 & 141.816 & 187.528 & 24.212 \\
 & MAC & 231,048 & 8 & 5,578,560 & 3,271,688 & 99,000 & 1,378,080 & 690,376 & 26,450,432 & 342,766 & 65,290 & 128,426 & 528 & 21,934,272 & 184,064 \\
 & Energy & 3.31E-06 & 1.75E-08 & 5.00E-03 & 4.30E-03 & 4.13E-04 & 2.64E-03 & 9.85E-04 & 7.98E-02 & 6.68E-04 & 4.14E-04 & 3.93E-04 & 3.50E-08 & 1.06E-02 & 2.55E-04 \\
 & Latency & 5.521 & 0.011 & 342.462 & 429.557 & 31.991 & 199.602 & 70.924 & 5128.256 & 71.745 & 31.542 & 42.347 & 0.057 & 677.695 & 26.393 \\ \midrule
\multirow{6}{*}{HouseholdPowerConsumption2} & RMSE & 50.930 & 172.072 & 50.086 & 54.395 & 57.065 & 53.406 & 64.535 & 184.172 & 54.538 & 59.565 & 54.566 & 55.732 & 55.729 & 52.349 \\
 & FLASH & 234.184 & 8.976 & 13.160 & 26.888 & 7.136 & 13.256 & 10.824 & 164.952 & 16.488 & 55.552 & 11.048 & 39.232 & 39.232 & 381.328 \\
 & RAM & 15.688 & 37.692 & 72.196 & 63.020 & 16.644 & 95.924 & 94.816 & 774.064 & 22.684 & 25.332 & 18.636 & 141.816 & 141.816 & 50.092 \\
 & MAC & 231,048 & 8 & 5,578,560 & 3,271,688 & 99,000 & 1,378,080 & 690,376 & 26,450,432 & 274,188 & 282,567 & 128,426 & 528 & 528 & 2,787,856 \\
 & Energy & 3.31E-06 & 1.75E-08 & 5.00E-03 & 4.30E-03 & 4.13E-04 & 2.64E-03 & 9.85E-04 & 7.98E-02 & 6.02E-04 & 5.98E-04 & 3.93E-04 & 3.50E-08 & 3.50E-08 & 2.85E-03 \\
 & Latency & 5.521 & 0.011 & 342.462 & 429.557 & 31.991 & 199.602 & 70.924 & 5128.256 & 60.644 & 52.580 & 42.347 & 0.057 & 0.057 & 242.887 \\ \midrule
\multirow{6}{*}{\emph{Average} $\downarrow$} & RMSE & 102.524 & 879.733 & 106.832 & 237.051 & 189.274 & 106.556 & 744.649 & 552.171 & 106.092 & 103.902 & 105.494 & 292.259 & 106.322 & \textbf{102.409} \\
 & FLASH & 234.184 & 8.976 & 13.160 & 26.888 & 7.136 & 13.256 & 10.824 & 167.520 & 17.308 & 33.768 & 11.048 & 39.232 & 1503.872 & 197.784 \\
 & RAM & 15.688 & 37.692 & 72.196 & 63.020 & 16.644 & 95.924 & 94.816 & 774.000 & 24.220 & 22.282 & 18.636 & 141.816 & 164.672 & 37.152 \\
 & MAC & 231,048 & 8 & 5,578,560 & 3,271,688 & 99,000 & 1,378,080 & 690,376 & 26,450,432 & 308,477 & 173,929 & 128,426 & 528 & 10,967,400 & 1,485,960 \\
 & Energy & 3.31E-06 & 1.75E-08 & 5.00E-03 & 4.30E-03 & 4.13E-04 & 2.64E-03 & 9.85E-04 & 7.98E-02 & 6.35E-04 & 5.06E-04 & 3.93E-04 & 3.50E-08 & 5.30E-03 & 1.55E-03 \\
 & Latency & 5.521 & 0.011 & 342.462 & 429.557 & 31.991 & 199.602 & 70.924 & 5128.256 & 66.194 & 42.061 & 42.347 & 0.057 & 338.876 & 134.640 \\ \bottomrule
\end{tabular}%
}
\caption{Full experimental results on \textbf{noisy} and \textbf{irregular time-series} classification and regression tasks, comparing downstream error (RMSE), accuracy, and model complexity metrics.} \label{table:irregular_results}
\end{table*}

\twocolumn
\section{Full Experimental Results} \label{section:full_results}

\add{In Tables \ref{table:classification} and \ref{table:regression}, we present the full results for classification and regression tasks, respectively. The model performance metric for classification is accuracy and for regression is RMSE. The model complexity metrics extensively include FLASH, RAM, MAC, Latency, and Energy Consumption. In addition, \autoref{table:irregular_results} presents the complete results for irregular and noisy time-series datasets.}

\section{\add{Resource Cost Comparison}} \label{section:cost_comparison}

\add{To quantify \algname{}’s relative efficiency, we compare the resource cost—specifically, the runtime—for the main results presented in Tables \ref{table:classification} and \ref{table:regression}. The results are reported in \autoref{table:cost_comparison}. For grid search and random search, we set the number of rounds and the number of candidates per round to be the same as in \algname{}, i.e., $B = 3$ and $C = 5$. Thus, we have 15 candidates in total and select the best one based on validation performance. For TinyTNAS, we set the permissible search time to be twice the average time of \algname{}, i.e., 8 minutes.}

\add{According to the results, \algname{} consistently achieves significantly lower search costs compared to traditional NAS methods such as grid search and random search. On average, \algname{} takes only 229.57 seconds (including image generation in the multimodal query generation preprocessing time), whereas grid search and random search take 1,356.78 and 1,713.18 seconds, respectively. Even compared to the efficient TinyTNAS, which averages 221.36 seconds, \algname{} performs comparably while maintaining or improving performance, especially across various classification and regression datasets. This result highlights the practical efficiency and scalability of \algname{}.}

\begin{table}[t]
\centering
\resizebox{\linewidth}{!}{%
\begin{tabular}{@{}ccccc@{}}
\toprule
\textbf{Datasets} & \textbf{Grid Search} & \textbf{Random Search} & \textbf{TinyTNAS} & \textbf{\algname{}} \\ \midrule \midrule
AppliancesEnergy & 1,195.44 & 691.18 & 151.72 & 332.53 \\
BIDMC32SpO2 & 489.21 & 605.25 & 156.31 & 178.19 \\
BenzeneConcentration & 1,156.71 & 718.81 & 424.27 & 324.53 \\
FloodModeling & 4,385.44 & 3,872.19 & 241.79 & 151.85 \\
LiveFuelMoistureContent & 347.17 & 1,891.60 & 265.96 & 175.51 \\ \midrule
AtrialFibrillation & 1,145.92 & 3,249.92 & 215.81 & 165.50 \\
BinaryHeartbeat & 155.91 & 185.56 & 217.73 & 170.40 \\
Cricket & 128.87 & 153.04 & 176.87 & 300.68 \\
FaultDetectionA & 1,018.39 & 2,190.53 & 122.50 & 216.64 \\
UCIHAR & 3,544.73 & 3,573.73 & 240.68 & 279.88 \\ \midrule
\emph{Average} & 1,356.78 & 1,713.18 & 221.36 & 229.57 \\ \bottomrule
\end{tabular}%
}
\caption{Search cost (in seconds) comparison between traditional NAS methods and \algname{}.} \label{table:cost_comparison}
\end{table}

\end{document}